\pdfoutput=1

\documentclass[11pt]{article}

\usepackage{acl}

\usepackage{times}
\usepackage{latexsym}
\usepackage{amsmath}
\usepackage{comment}
\usepackage{tabularx}
\usepackage{multirow}
\usepackage{rotating}
\usepackage{placeins}
\usepackage{booktabs}
\usepackage{enumitem}
\usepackage{mleftright}
\usepackage{amssymb}
\usepackage{nicefrac}
\usepackage{caption}
\usepackage{subcaption}

\usepackage[T1]{fontenc}

\usepackage[utf8]{inputenc}

\usepackage{microtype}

\usepackage{inconsolata}

\usepackage{graphicx}

\usepackage{natbib}

\usepackage[printonlyused,nolist,nohyperlinks]{acronym}

\newcommand\blfootnote[1]{%
  \begingroup
  \renewcommand\thefootnote{}\footnote{#1}%
  \addtocounter{footnote}{-1}%
  \endgroup
}

\newcommand{\nhphantom}[1]{\sbox0{#1}\hspace{-\the\wd0}}

\title{
From Argumentation to Deliberation: Perspectivized Stance Vectors\\ for Fine-grained (Dis)agreement Analysis
}

\author{
\\
\textbf{Moritz Plenz}$^{\clubsuit\heartsuit}$\ \ ~~\textbf{Philipp Heinisch}$^{\spadesuit\heartsuit}$\ \ ~~\textbf{Janosch Gehring}$^{\clubsuit}$\\ \textbf{Philipp Cimiano}$^{\spadesuit}$\ \ ~~\textbf{Anette Frank}$^{\clubsuit}$ \\\\
$^{\clubsuit}$\,Heidelberg University %
\ \ ~~$^{\spadesuit}$\,Bielefeld University \\
\texttt{\{plenz,gehring,frank\}@cl.uni-heidelberg.de}\ \ \ \ \ \ \ \\
\ \ \ \ \ \ \texttt{\{pheinisch,cimiano\}@techfak.uni-bielefeld.de}
}

\begin{document}
\maketitle
\begin{abstract}

Debating over conflicting issues is a necessary first step towards resolving conflicts. However, intrinsic perspectives of an arguer are difficult to overcome by persuasive argumentation skills. Proceeding from a debate to a \textit{deliberative process}, where we can identify actionable options for resolving a conflict requires a deeper analysis of arguments and the perspectives they are grounded in -- as it is only from there that one can derive  mutually agreeable resolution steps. In this work we develop a framework for a \textit{deliberative analysis of arguments} in a computational argumentation setup. We conduct a fine-grained analysis of \textit{perspectivized stances} expressed in the arguments of different arguers or stakeholders on a given issue, aiming not only to identify their opposing views, but also shared perspectives arising from 
their attitudes, values or needs.
We formalize this analysis in \textit{Perspectivized Stance Vectors} that characterize the individual perspectivized stances of all arguers on a given issue. We construct these vectors by determining \textit{issue}- and \textit{argument-specific concepts}, and predict an arguer's 
stance relative to each of them. The vectors allow us to measure a \textit{modulated (dis)agreement} between arguers, structured by perspectives, which allows us to identify actionable points for conflict resolution, as a first step towards deliberation.\blfootnote{\nhphantom{$^\heartsuit$}$^\heartsuit$Authors contributed equally.}\footnote{Data and code are available at \\ \url{https://github.com/Heidelberg-NLP/PSV}.}
\end{abstract}

\section{Introduction}

Diverse stakeholders exchange their opinions and arguments on social media, news, debating portals and other private or public discussion formats. Often, they are in strong opposition, leaving little room for a consensus that could resolve the conflict.  
While argument mining technology has concentrated on analysing and generating arguments that can support arguers in winning a debate \citep{habernal-gurevych-2016-makes,wang-etal-2017-winning,wachsmuth-etal-2018-retrieval}, so far there has been limited interest in identifying points in opposing positions that bear a chance for consensual resolution of the conflict. 
Identifying points that offer a chance for resolution requires fine-grained analysis of the stances expressed by different stakeholders, to understand on which specific aspects they disagree and on which they actually might agree, and which of these are crucial for their mutual (dis)agreement. 

\begin{figure}
    \centering
     
     \includegraphics[width=1.0\linewidth]{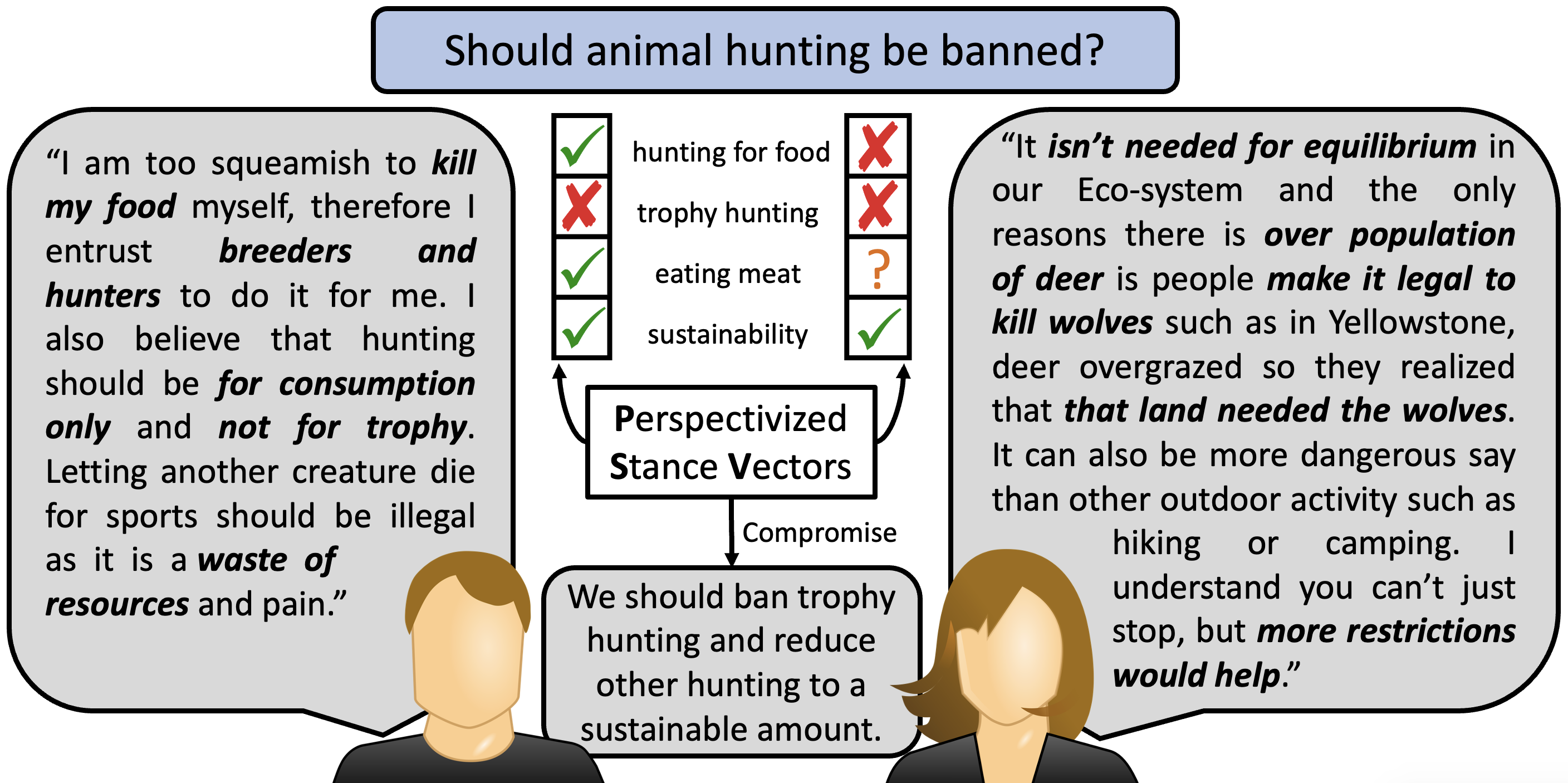}
    \caption{Example PSVs for `\textit{Animal Hunting}'.}
    \label{fig:animal-hunting-PSVs}
\end{figure}

This requires an analysis of the \textit{perspectives} an arguer has on an issue -- which may be grounded in their values, attitudes or specific goals and needs \citep{falk-etal-2024-overview,kiesel-etal-2022-identifying,alshomary-etal-2022-moral}. 
We aim to compare arguments within a given issue based on their expressed perspectives, which means that we require a fixed set of perspectives for each issue. Issue-specific `frames' are commonly used to group and analyze arguments from a given issue \citep{opitz-etal-2021-explainable,heinischetal:2022a}, which makes them promising for modeling perspectives. Following \citet{plenz-etal-2024-pakt}, we use a data-driven approach to extract issue-specific frames from the commonsense knowledge graph ConceptNet \citep{speer2017conceptnet}, meaning that \textit{concepts} (i.e., nodes from ConceptNet) form our basis for perspectives. %
To support a \textit{deliberative analysis of arguments}, we develop tools to i) determine \textit{relevant concepts} that characterize different perspectives arguers may have on an issue and ii) what \textit{stance} arguers express towards a certain perspective with a given argument. 
Our rationale is that by determining on which specific perspectives arguers \textit{agree} or \textit{disagree} in a debate, one may be able to identify points for achieving consensual agreement. In the following we thus use the terms \textit{perspective} and \textit{concept} interchangeably, although we want to note that perspectives may be formalized differently in future work. 

Towards this goal, our work presents a new approach to construct a fine-grained representation of arguments that characterizes the \textit{perspectivized stances} arguers express on a given issue, in so-called \textit{Perspectivized Stance Vectors} (PSVs). A PSV is formalized as a vector of \textit{stance values} towards issue-specific concepts (perspectives) -- the so-called \textit{signature}. 
Comparing multiple PSVs can reflect \textit{opposing}, \textit{agreeing} and \textit{orthogonal} stances of \textit{different strengths} for \textit{different perspectives}, offering ways to identify potential anchors for deliberation processes. This goes beyond conventional stance classification, which only allows to identify conflicts at a binary level -- instead, our analysis allows for more fine-grained assessments. 

Figure~\ref{fig:animal-hunting-PSVs} illustrates the concept of PSVs applied to two arguments on the issue ``\textit{Should animal hunting be banned?}'' 
Choosing four example key perspectives from the debate, the arguer on the left is clearly \textsc{pro} \textit{hunting for food}, \textit{sustainability} and \textit{eating meat}, but \textsc{against} \textit{trophy hunting}. 
The position of the arguer can thus be represented as a 4-dimensional vector, where the dimensions correspond to the above-mentioned perspectives. 
The vector for the second arguer will instead be: \textsc{pro}: \textit{sustainability}; \textsc{against}: \textit{hunting for food}, \textit{trophy hunting}; and \textsc{neutral}: \textit{eating meat}.

The vectors show opposing stance on \textit{hunting for food}, but 
agreement for \textit{trophy hunting} (\textsc{against}) and \textit{sustainability} (\textsc{pro}). \textit{Eating meat} does not show agreement nor disagreement, and hence is considered orthogonal. 
Agreeing dimensions could offer an entry point and basis for resolving the conflict by emphasizing shared positions, and aiming to find consensual solutions on points of opposite stances. Hence, ``\textit{We should ban trophy hunting and reduce other hunting to a sustainable amount}'' could be a viable resolution.

\begin{enumerate}[label={\roman*)}, noitemsep, leftmargin=*]
\item[] \hspace*{-0.5cm} In summary, our contributions are:

\item We formalize \textbf{P}erspectivized \textbf{S}tance \textbf{V}ectors (PSVs) as a structured representation of arguments to enable a \textit{deliberative analysis}. 
\item This includes three subtasks: To construct PSVs we need to i) select issue-specific \textit{signature concepts} and ii) classify the corresponding perspectivized \textit{stances values} for a given argument. To identify (dis)agreement we need to iii) \textit{aggregate PSVs}. 
\item We run experiments on deliberative issues from \textsc{pakt} \citep[a structured argumentation corpus;][]{plenz-etal-2024-pakt} and compare to a manually annotated evaluation set, showing our individual modules' performances. %
\item Our evaluations include case studies as proof-of-concept on how a perspectivized analysis of
acceptability can support deliberation.  
\end{enumerate}

\section{Related Work}
\paragraph{Overall stance.} 
PSVs are a fine-grained representation of stances. Classifying the overall stance of arguments towards a topic is a core task in argument mining~\citep{bar-haim-etal-2017-stance,kobbe-etal-2020-unsupervised,luo-etal-2020-detecting}. 
However, assigning an argument a global stance tag (e.g., \textsc{pro}, \textsc{con} and \textsc{neutral} or \textsc{unrelated}) lacks expressivity: it divides sets of arguments into only a couple of groups, neglecting crucial nuances. The task of same-side classification (predict whether two arguments share their overall stance) in \citet{hou-jochim-2017-argument,korner-etal-2021-classifying} does not address this problem either. Further, it does not unveil the underlying reasons why arguments share a stance. 

To counter this issue, prior work incorporated background knowledge, by including \textit{reasoning paths} to \textit{explain}, e.g., for which reasons a premise supports or attacks a conclusion \citep{pauletal:2020}, or to generate an \textit{explanation graph} for a premise-conclusion pair that explains the stance of the argument \citep{saha-etal-2021-explagraphs,saadat-yazdi-etal-2023-uncovering,plenz-etal-2023-similarity}. We build on this work by including concepts from a commonsense resource to define the PSV signature concepts. %

\paragraph{Perspectives in argumentation.}

Our work is related to \citet{barrow-etal-2021-syntopical}, who rely on graphs to represent arguments and their relationships as a basis to detect viewpoints. They proposed so-called \textit{syntopical graphs} that model pairwise textual relationships between claims to enable a better reconstruction of latent viewpoints in a collection, thereby making points of (dis)agreement within the collection explicit. %
In a similar way, PSVs enable the detection of (dis)agreement. But in addition, %
PSVs can detect \textit{orthogonality}, i.e., cases where a pair of arguments is not 
related to each other. 

Our work is also related to the analysis of framing in argumentation 
\citep{heinischetal:2022a,heinisch-etal-2023-unsupervised,otmakhova-etal-2024-media}, where emphasized aspects are automatically detected. %
Specifically, our work is related to the idea that frames %
can be issue- or topic-specific and thus need to be identified in a bottom-up fashion for each topic. 
\citet{ajjour-etal-2019-modeling} present an unsupervised approach that induces frames by clustering arguments from an issue. 
\citet{ruckdeschel-wiedemann-2022-boundary}, by contrast, present a topic-specific framing approach, with the limitation of training a classifier for each topic separately -- which then cannot be applied to new topics. 
Our unsupervised approach to signature induction follows \citet{plenz-etal-2024-pakt}, who find that the knowledge base ConceptNet~\citep{speer2017conceptnet} provides suitable, often implicit concepts relevant for argument interpretation. %

Finally, our task is related to aspect-based sentiment analysis \citep[ABSA; ][]{cabello-Akujuobi-2024-it,wang-etal-2024-refining,frasincar_2023,hoang-etal-2019-aspect}, which aims for a fine-grained view on which aspects are target of a certain sentiment. 
While ABSA is typically applied to reviews, we aim for a fine-grained, perspectivized analysis of arguments in deliberation, by detecting argument-related stances towards specific concepts. 

Our framework supporting a deliberative analysis of arguments thus brings together and combines methods from viewpoint detection, framing and aspect-based sentiment analysis. We combine these methods in a novel way for deliberation support, by pinpointing conflicting perspectives and concepts between argument and stakeholders, with the aim of resolving conflicts and suggesting compromises. 

\paragraph{Deliberation} 
refers to a collaborative argumentative exchange where arguers hold incompatible views on an issue, which they seek to resolve by achieving a consensual decision \citep{FELTON2022100350}. Deliberative processes are naturally framed as a dialogue \citep{waltonetal-2010-deliberation,snaith-etal-2010-mixed,Waltonetal-2016-deliberation}. %
For example, \citet{al-khatib-etal-2018-modeling} successfully classify different \textit{strategies} of participants
in deliberative discussions. Yet they do not evaluate the underlying perspectives, nor the effectiveness of these strategies for the aim of achieving an agreeable resolution of conflicts.

Deliberation is typically approached using pre\-fer\-ence frameworks that take into account the arguers' diverging desires or goals, or their normative or moral considerations \citep{modgil,Amgound-etal-2014-preference,10.1007/978-3-319-69131-2_25}. We do not focus on algorithmic resolution of conflicts, but on analyzing the arguers' perspectivized viewpoints to quantify dimensions of (dis)agreement -- which future work may extend with reasoning processes, to derive potential resolutions. 

\citet{RECAP-2018} provide overviews of debates to make decision makers aware of arguments and opinions on relevant topics. Using a Case-Based Reasoning approach, they compute similarity between arguments to retrieve or cluster similar arguments. This allows them to synthesize new arguments -- by extrapolating from and combining existing arguments. 
While they focus on grouping similar arguments, we aim for an aggregated representation of debates in terms of perspectivized stances that reflect diverging and unified viewpoints of relevant stakeholder groups.

Some recent work leverages LLMs to model deliberative processes. E.g., \citet{bakker-etal-2022-finetuning} investigate if LMs can support humans in finding agreement on conflicting issues. They task LLMs to expand a corpus from a set of human-elicited questions and opinions on moral and political issues, 
and train a reward model to rate LM-generated consensus statements. 
They report high performance of LLMs generating consensual statements. However the evaluations do not report detailed statistics, and since the data
(worth £46,000) is not made public, it remains unclear if the evaluation involves notable conflicts to start with.\footnote{For evaluation,  opinions are clustered by topic (\textit{not} by the original issues). 
Fig.\ 2B of the work splits the data into divisive and non-divisive questions within a group, where only 50\% were found to be divisive. The win rates for their model over baselines are similar between (non)divisive  questions, and the analysis does not detail agreement score differences between positioned vs.\ agreement statements for the divisive subset. Without access to the data nor detailed analyses, it is unclear whether the re-clustered data involves notable conflicts.}

Our work is of smaller scale, but relies on arguments from a curated and accessible debate portal. 
In contrast to their work -- which is elusive on %
\textit{which} divisive arguments a consensus statement is meant to resolve --
we explicitly represent arguments as stance vectors along conceptual perspectives, from which we compute highly interpretable acceptability scores as a basis for finding consensual solutions for conflicting arguments. Since our method is interpretable, this can 
increase trust, and thereby usability, compared to purely generative methods.

\section{Perspectivized Stance Vectors}

We introduce \textbf{P}erspectivized \textbf{S}tance \textbf{V}ectors (PSVs), a new representation to record the perspectives expressed in or underlying an argument, with the aim to detect and measure agreement and conflicts between pairs or a set of arguments on a given issue. To construct a PSV, we need to define its \emph{signature} and corresponding \emph{stance values}. Given a debate topic $dt$, the signature is determined by a list of concepts $\{c_i\}_{i=1}^n$ relevant for topic $dt$. 

Given an argument $a$ on topic $dt$, we abstract a PSV $\vec{v_a}$ from argument $a$ by determining a stance $s_i$ for each concept $c_i$, where $s_i$ represents the stance the arguer expresses with argument $a$ towards the concept $c_i$. We choose stance values $s_i \in [-1, 0, 1]$ to represent a stance of \textit{against}, \textit{neutral}, or \textit{in favor}, respectively. %
We formalize PSVs as either $n$-dimensional vectors of stance values $s_i$, or $(n \times 3)$ dimensional matrices where each entry represents the probability of concept $c_i$ to belong to one of the three stance value classes: 
\begin{equation}
\begin{aligned}
    \vec{s} &\in \mathbb{S}^n \quad &\mathrm{where} \quad \mathbb{S} &= \{-1, 0, 1\} \\
    \vec{p} &\in \mathbb{P}^{n \times 3} \quad &\mathrm{where} \quad \mathbb{P} &= [0, 1].
\end{aligned}
\end{equation}
If the exact representation is not relevant, we use $\vec{v}$ to denote a general PSV. 
When comparing pairs of PSVs ($\vec{v}_{a_1}$, $\vec{v}_{a_2}$) for arguments ($a_1$, $a_2$), %
aligned vs.\ opposing dimensions indicate agreement or disagreement, respectively. Dimensions with neutral stance labels in ($\vec{v}_{a_1}$, $\vec{v}_{a_2}$) indicate \textit{orthogonality}, as the arguers neither agree nor disagree. 

We next describe our methods to construct PSVs, including their signatures and stance values in \S\ref{sec:create-psv}. We then describe how to aggregate or compare PSVs to obtain predictions for agreement, orthogonality and disagreement between arguments and which specific perspectives cause them (\S\ref{sec:method:acceptability_scores}). 

\subsection{PSV Construction} \label{sec:create-psv}

Below we show how to construct PSVs given a topic $dt$ and a set of arguments $\mathcal{A}_{dt}$ on that topic. 

\subsubsection{\textit{Signature concepts} for Debate Topic} \label{sec:method:signature_concepts}
As a signature for PSVs we are interested in general -- and potentially conflicting -- concepts that capture the perspectives of diverse arguers towards a topic. 

Following \citet{plenz-etal-2023-similarity,plenz-etal-2024-pakt}, we align arguments to the commonsense knowledge graph ConceptNet~\citep{speer2017conceptnet}. First, we split arguments into individual sentences, then we select for each sentence the most similar concepts (i.e., nodes in ConceptNet). 
We connect these concepts with weighted shortest paths that maximize semantic similarity to the argument. Concepts along such paths have been 
shown to cover relevant aspects of the given text, while maintaining high precision~\citep{plenz-etal-2023-similarity,fu-frank-2024-mystery}. These nodes form a set of concepts $C_a$ that reflects the given argument $a \in \mathcal{A}_{dt}$. 

To obtain conflicting concepts, we split the arguments $\mathcal{A}_{dt}$ by their overall stance towards the topic into $\mathcal{A}_{dt}^+$ and $\mathcal{A}_{dt}^-$. For each concept $c_i \in \cup_{a\in\mathcal{A}_{dt}}C_a$ and debate topic stance $s_{dt}\in\{+,-\}$ we compute the stance-specific frequency 
\begin{equation}
f_{c_i}^{s_{dt}} = \frac{\# \{a \,|\, a \in \mathcal{A}_{dt}^{s_{dt}} \ \mathrm{and}\  c_i \in C_a\}}{\mathrm{max}\left(1, \, \# \{a \,|\, a \in \mathcal{A}_{dt}^{s_{dt}}\}\right)}. 
\end{equation}
We normalize the frequencies $f_{c_i}^{s_{dt}}$ for a given concept and stance by subtracting the frequencies of the concept with the opposite stance: $f_{c_i}^+ - f_{c_i}^-$ and $f_{c_i}^- - f_{c_i}^+$ for \textsc{pro} and \textsc{con} stance, respectively. 
To avoid redundancy, we remove concepts with duplicate lemmas. Finally, we take the top-$k$ \textsc{pro} and \textsc{con} concepts, resulting in $2\cdot k$ concepts in total. 

Optionally, the resulting concepts can be filtered to obtain smaller and more concise PSVs: 
we either remove hypernyms of other concepts in order to further \textit{reduce redundancy}, or remove concepts that ChatGPT judges as irrelevant for the topic, to \textit{increase relevancy}. Refer to \S\ref{app:sec:signature} for more details. 

Ideally, the signature concepts are \textit{relevant} to the debate topic and of appropriate \textit{granularity}: If a signature concept is too fine-grained, then only very few (or no) arguments will evoke it. Similarly, if a signature concept is too coarse-grained then arguments may not have a clear stance towards it -- consider for example the concept ``hunting'' for the first argument in Fig.~\ref{fig:animal-hunting-PSVs}. Here, ``hunting for food'' and ``trophy hunting'' would form a better signature. Hence, we aim for signature concepts with an intermediate granularity. In our experiments (\S\ref{sec:exp:construction}) we asses the (i) \textit{relevance} to the debate topic and (ii) \textit{granularity} of selected signature concepts. 

\begin{table*} %
    \centering
    \resizebox{\linewidth}{!}{
    \begin{tabular}{llccc}
    \toprule
        \multicolumn{2}{l}{Method} & Agreement [$+$] & Orthogonal [$\varnothing$] & Disagreement [$-$] \\
    \cmidrule(r){1-2} \cmidrule(l){3-5}
        $\mathcal{S}$ & Stance Value & 
            $\delta\mleft(s^1_i, s^2_i\mright)$ & 
            -- & 
            $1 - \delta\mleft(s^1_i, s^2_i\mright)$ \\[5pt]
        $\mathcal{S}_0$ & Stance Value (Consid. Neut.) & 
            $\delta\mleft(s^1_i, s^2_i\mright) \left(1 - \delta\mleft(s^1_i, 0\mright)\right)$ & 
            $\mathrm{min}\mleft(\sum\nolimits_{j=1}^2\delta\mleft(s^j_i, 0\mright), 1\mright)$ & 
            $\left(1 - \delta\mleft(s^1_i, s^2_i\mright)\right) \left(1 - \delta\mleft(s^1_i, 0\mright)\right) \left(1 - \delta\mleft(s^2_i, 0\mright)\right)$ \\[5pt]
        $\mathcal{S}_D$ & Stance Value (Difference) & 
            $\mathcal{S}_0^+\mleft(s_i^1, s_i^2\mright) - \mathcal{S}_0^-\mleft(s_i^1, s_i^2\mright)$ &
            -- & 
            $\mathcal{S}_0^-\mleft(s_i^1, s_i^2\mright) - \mathcal{S}_0^+\mleft(s_i^1, s_i^2\mright)$ \\
        \cmidrule(r){1-2} \cmidrule(l){3-5}
        $\mathcal{P}$ & Stance Prob. & 
            $\left(p^1_i \odot p^2_i\right) \cdot [1,1,1]^T$ &
            -- & 
            $\nicefrac{1}{2} \left|p_i^1 - p_i^2\right| \cdot [1,1,1]^T$ \\[5pt]
        $\mathcal{P}_0$ & Stance Prob. (Consid. Neut.) &
            $\left(p^1_i \odot p^2_i\right) \cdot [1,0,1]^T$ &
            $\left(p^1_i \odot p^2_i\right) \cdot [0,1,0]^T$ & 
            $\nicefrac{1}{2} \left|p_i^1 - p_i^2\right| \cdot [1,0,1]^T$ \\[5pt]
        $\mathcal{P}_D$ & Stance Prob. (Difference) & 
            $\mathcal{P}_0^+\mleft(p_i^1, p_i^2\mright) - \mathcal{P}_0^-\mleft(p_i^1, p_i^2\mright)$ &
            -- & 
            $\mathcal{P}_0^-\mleft(p_i^1, p_i^2\mright) - \mathcal{P}_0^+\mleft(p_i^1, p_i^2\mright)$ \\
    \bottomrule
    \end{tabular}
    }
    \caption{Aggregation Methods. $s_i^j \in \{-1, 0, 1\}$ is the stance value of argument $j$ towards concept $i$ and $p_i^j \in [0,1]^{(1\times3)}$ are the corresponding probabilities. The Kronecker delta $\delta\mleft(x,y\mright)$ is 1 if $x=y$ and 0 else. $\odot$ is element-wise multiplication. \S\ref{app:sec:method:aggregation_methods} discusses the formulas in more detail. }
    \label{tab:aggregation_methods}
\end{table*}

\subsubsection{\textit{Perspectivized stances} for Arguments} \label{sec:method:perspectivized_stances}
We develop different models to compute the stance value $s_i$ for a given argument and signature concept $c_i$. Here we provide an overview of the methods, please refer to \S\ref{app:sec:stance_values} for more in-depth descriptions. 

\paragraph{Baseline.} 
Our Baseline assigns each argument $a$ and concept $c_i$ the stance value $0$ if the concept is \textit{not} in the concept graph, i.e., $s_i = 0 \ \mathrm{for} \ c_i \not\in C_a$. Else, for concepts that are in the argument graph $C_a$, the \textit{debate topic stance} $s_{dt}\in\{+1,-1\}$ is assigned for concept $c_i$. 

\paragraph{\texttt{RoBERTa}.}
We construct a synthetic dataset by automatically adapting %
sentiment~\citep{SobhaniMK16,MohammadSK17} and human-value~\citep{mirzakhmedova-etal-2024-touche23-valueeval} detection datasets to our task. On this data we apply \textit{transfer learning} using a RoBERTa model that has been fine-tuned for sentiment analysis~\citep{loureiro-etal-2022-timelms}, by further training it on our synthetic dataset. We emphasize that this synthetic data is exclusively used to finetune \texttt{RoBERTa}, which is not our best-performing system -- and our remaining  experiments are independent of this synthetic data. We refer to appendix \ref{app:sec:stance_values} for more details. %

\paragraph{\texttt{GPT4o}.} We also prompt \texttt{GPT4o} to predict whether an argument $a$ is against, neutral or in favor of a concept $c_i$. We apply zero- and few-shot prompting, with two hand-crafted samples for the latter (\ref{app:sec:stance_values}).

\subsection{Computing \textit{Acceptability} Scores} 
\label{sec:method:acceptability_scores}
Standard stance classification allows us to predict whether two arguments agree or disagree on a debate topic. 
Using PSVs, we can now detect and predict agreement on a more fine-grained level. E.g., there is a \textit{partial agreement} between the arguers in Fig.~\ref{fig:animal-hunting-PSVs}: Both parties are against \textit{trophy} hunting while they are in favor of \textit{sustainability}. 
Such \textit{partial agreements} are instrumental to find compromises between arguers who \textit{disagree} on a topic. 

\paragraph{\textit{Perspectivized Acceptability} Scores.} 
Our hypothesis is that i) arguers \textit{agree} or \textit{disagree} on the concepts $c_i$, depending on whether %
their arguments express %
the same perspectivized stances $s_i$ towards $c_i$ within the debated topic. %
Yet, ii) if at least one of two %
arguments has a neutral perspective towards concept $c_i$, then the arguers neither agree nor disagree on the concept, meaning they are \textit{orthogonal}. %
Following this intuition we design several aggregation methods to detect perspectivized \textit{agreement}, \textit{orthogonality} and \textit{disagreement} between arguments at concept level. 

The aggregation functions are shown in Table~\ref{tab:aggregation_methods}: we group them depending on whether they use discrete stance values ($\mathbb{S}=\{-1, 0, 1\}$) or continuous probabilities for the respective classes ($\mathbb{P}=[0,1]$). The functions return discrete or continuous predictions, respectively. Further, the aggregation can consider the special role of  neutral stance values, as outlined above. Finally, for agreement and disagreement, we can consider the opposing class
as an inhibiting factor. Hence, we also design 
functions that take the difference between agreement and disagreement. We refer to these functions as, e.g., $\mathcal{S}^-_0\mleft(s_i^1, s_i^2\mright)$ to denote a disagreement score (superscript ``$-$'') between arguments $a_1$ and $a_2$ regarding concept $i$ (function parameter) under a stance value-based aggregation function ($\mathcal{S}$) that considers the role of neutral stances (subscript $0$). 

For example, $\mathcal{S}^+\mleft(s_i^1, s_i^2\mright) = 1$ iff the stance values are equal $s_i^1=s_i^2$, while $\mathcal{S}_0^+\mleft(s_i^1, s_i^2\mright) = 1$ and $\mathcal{S}_D^+\mleft(s_i^1, s_i^2\mright)=1$ iff the stance values are equal and non-zero $s_i^1=s_i^2 \in \{-1, +1\}$. Compared to $\mathcal{S}_0^+\mleft(s_i^1, s_i^2\mright)$, $\mathcal{S}_D^+\mleft(s_i^1, s_i^2\mright)$ also distinguishes between agreement scores of $0$ and $-1$. 
\S\ref{app:sec:method:aggregation_methods} explains the functions in more detail.

\paragraph{Acceptability scores between pairs of arguments.}
By now we discussed how to calculate contributions of individual perspectives towards agreement, orthogonality or disagreement. To obtain an overall acceptability score for an argument pair, we average perspectivized stance values of all dimensions $n$ of a PSV. While future work could investigate the effects of making the contributions of each perspective learnable, for the scope of this paper we restrict ourselves to unsupervised aggregation methods. 

\section{
    PSV-Experiments
}

Given a set of arguments on an issue, our approach first finds signature concepts, then computes perspectivized stances which yields PSVs and finally aggregates PSVs to obtain acceptability scores. In this section, we empirically assess the quality of each of these steps by comparing to human annotations. Where possible, we augment our manual evaluation with automatic evaluations that do not require human labels. Section~\ref{sec:case_study} presents a complementary case study.

\subsection{Experimental setup} \label{sec:exp:setup}
\paragraph{Data.}
We conduct our analyses and case study using \textsc{pakt}~\citep{plenz-etal-2024-pakt}, a debate resource that presents issues as binary questions, and answers to these questions as arguments for either stance. 
The arguments, on avg.\ 7 sentences long, discuss
an author's points without elaborating on the entire issue. 
This makes \textsc{pakt} well-suited for our purposes. Fig.~\ref{fig:animal-hunting-PSVs} shows two shortened example arguments from \textsc{pakt}. For our case study we further enrich \textsc{pakt} with stakeholder groups (for details see \S\ref{app:sec:stakeholder}). 

\paragraph{Annotation.}  %
To assess the quality of our methods, %
three annotators labeled data from 5 different evenly represented topics: 300 topic-level annotations to evaluate PSV signatures, 500 argument-level annotations to evaluate PSV values and 1,500 annotations for pairs of arguments to evaluate our methods to predict acceptability scores on debate topics. We collect annotations from all three annotators for the topic \textit{Should animal hunting be banned?} to estimate inter-annotator agreement. For most subtasks we achieve moderate to high agreement as shown by Tab.~\ref{app:tab:annotation}, despite the high subjectivity in argumentative tasks. \S\ref{app:sec:annotation} presents more details on the annotation procedure. 

\subsection{Analyzing PSV Construction Methods} \label{sec:exp:construction}

First, we analyze how best to construct PSVs. This includes i) the selection of perspectives for the PSV signature and ii) how to predict 
PSV values. 

\paragraph{PSV signature.} Signature perspectives should be \textit{relevant} to the topic. %
Also, if a perspective is too \textit{general} (or  too \textit{specific}), it will be evoked by almost all (or no) arguments. Neither is useful for comparing arguments -- hence, we %
check whether our signature concepts' granularity is in-between. 

Tab.\ \ref{tab:psv_signature} presents the \textit{relevance} and \textit{granularity} scores evaluated against our human annotation. 

\textit{Relevance.} Unfiltered perspectives (\textit{all}) are mostly relevant (Prec: 90.0\%). This can be increased to 93.8\% by filtering for irrelevance with ChatGPT -- at the cost of discarding other relevant perspectives (Rec: 55.6\%). The unfiltered approach yields the maximum F1-score of 94.7\%. 

\textit{Granularity.} Without filtering (\textit{all}) only 53.3\% of perspectives have appropriate granularity. Filtering for irrelevance raises precision to 75.0\%, while discarding some appropriate perspectives (Rec: 75.0\%), and yields the overall highest granularity F1-score of 75.0\%. 
This indicates that no filtering or irrelevance-filtering should yield best performances for granularity, depending on whether recall or precision is more important. Note, however,
that hypernym filtering %
aims to reduce redundancies in the selected perspectives, %
which we do not assess 
in our human annotation. Hence, hypernym filtering might perform well in practice although our human annotation does not capture its advantages.

\begin{table}[t]
    \centering
    \resizebox{0.80\linewidth}{!}{
        \begin{tabular}{@{}llrrrr@{}}
        \toprule
         &  & all & -hyp. & -irrel. & -both \\
        \cmidrule(r){1-2}\cmidrule(l){3-6}
        Avg & \# & 30.0 & 22.0 & 16.0 & 11.0 \\
        \cmidrule(r){1-2}\cmidrule(l){3-6}
        
         & P & 90.0 & 90.9 & \textbf{93.8} & 90.9 \\
        Relevance & R & \textbf{100.0} & 74.1 & 55.6 & 37.0 \\
         & F1 & \textbf{94.7} & 81.6 & 69.8 & 52.6 \\
        \cmidrule(r){1-2}\cmidrule(l){3-6}
        & P & 53.3 & 54.5 & \textbf{75.0} & 72.7 \\
        Granularity & R & \textbf{100.0} & 75.0 & 75.0 & 50.0 \\
         & F1 & 69.6 & 63.2 & \textbf{75.0} & 59.3 \\
        \bottomrule
        \end{tabular}
    }
    \caption{Analysis of Signature Perspectives, evaluated against human annotation of the \textit{Relevance} and appropriate \textit{Granularity} of concepts. %
    We present the unfiltered selection (\textit{all}), as well as three filtering methods: \textit{hyp} (hypernym) filtering, ChatGPT-based filtering for \textit{irrel}evance, and a combination of \textit{both}. \textit{Avg} shows the effects on the avg.\ nb.\ of perspectives (PSV dimension). We measure \textit{P}recision, \textit{R}ecall and \textit{F1}-score. %
    \label{tab:psv_signature}
    }
\end{table}

\paragraph{PSV stance values.} 
Table \ref{tab:psv_values} shows the performance of our PSV stance prediction methods compared to our annotations. \texttt{GPT4o} (zero-shot) is the best approach with 50.2\% macro F1 across all perspectives. With few-shot prompting the performance reduces to 49.0\%, but still outperforms our baseline and fine-tuned models. Overall the performances increase when considering only perspectives with appropriate granularity. These perspectives are less ambiguous and hence easier to annotate. Again, the \texttt{GPT4o}-based methods perform the best, achieving 56.3\% and 56.4\% macro F1 for zero- and few-shot, respectively. 

For our remaining analyses
we will use \texttt{GPT4o} (zero-shot) because of its overall best performance and simplicity. The confusion matrix (cf.\ Fig.~\ref{app:fig:confusion_matrix_stance_value_gpt4o} in \S\ref{app:sec:exp:stance_value}) reveals that \texttt{GPT4o} (zero-shot) performs well for positive and negative perspectives, but often misclassifies neutral ones as negative.

\begin{table}[ht]
    \centering
    \resizebox{\linewidth}{!}{
    \begin{tabular}{lrrrrr}
        \toprule
        Method & all & appropriate & negative & neutral & positive \\
        \cmidrule(r){1-1}\cmidrule(l){2-3}\cmidrule(l){4-6}
        Baseline & 38.4 & 41.3 & 20.1 & \textbf{71.0} & 24.1 \\
        1-\texttt{RoBERTa} & 36.5 & 37.9 & 45.5 & 28.5 & 35.4 \\
        3-\texttt{RoBERTa} & 43.2 & 43.9 & \textbf{46.6} & 47.0 & 35.9 \\
        \texttt{GPT4o} (0-shot) & \textbf{50.2} & 56.3 & 46.0 & 44.7 & \textbf{59.9} \\
        \texttt{GPT4o} (few-shot) & 49.0 & \textbf{56.4} & 45.3 & 43.0 & 58.6 \\
        \bottomrule
    \end{tabular}
    }
    \caption{PSV stance prediction. Scores are macro F1, evaluated on \textit{all} perspectives, or on perspectives with \textit{appropriate} granularity (c.f.\ \S\ref{sec:method:signature_concepts}). %
    We also report F1 scores for individual classes across all annotated perspectives.}
    \label{tab:psv_values}
\end{table}

\subsection{Evaluating Aggregation Methods}
Unless stated otherwise, the reported results use 100 dimensional PSVs w/o filtering and \texttt{GPT4o} (0-shot) for perspectivized stance prediction.

\subsubsection{Global acceptability: \textit{Partial agreement} among argument pairs of \textit{opposite stance}} \label{sec:exp:partial_agreement}

\begin{table}
    \centering
    \resizebox{0.85\linewidth}{!}{
    \begin{tabular}{clrrr}
        \toprule
        \multicolumn{2}{c}{Mode} & Agreement & Orthogonal & Disagreement \\
        \cmidrule(r){1-2} \cmidrule(l){3-5}
        \parbox[t]{2mm}{\multirow{6}{*}{\rotatebox[origin=c]{90}{Global}}}
        & $\mathcal{S}$ & 0.59 & -- & 0.67 \\
        & $\mathcal{S}_0$   & 0.55 & 0.66 & 0.68 \\
        & $\mathcal{S}_D$ & \textbf{0.60} & --   & 0.64 \\
        & $\mathcal{P}$ & 0.57 & -- & 0.67 \\
        & $\mathcal{P}_0$   & 0.54 & \textbf{0.69} & \textbf{0.70} \\
        & $\mathcal{P}_D$ & 0.58 & --   & 0.62 \\
        \cmidrule(r){1-2} \cmidrule(l){3-5}
        \parbox[t]{2mm}{\multirow{7}{*}{\rotatebox[origin=c]{90}{Perspectivized}}}
        & $\mathcal{S}$ & 0.54 & -- & 0.63 \\
        & $\mathcal{S}_0$   & 0.58 & 0.64   & 0.70 \\
        & $\mathcal{S}_D$ & 0.57 & --   & 0.67 \\
        & $\mathcal{P}$ & 0.52 & -- & 0.73 \\
        & $\mathcal{P}_0$   & \textbf{0.62} & \textbf{0.75}   & \textbf{0.76} \\
        & $\mathcal{P}_D$ & 0.56 & --   & 0.72 \\
        & w/o PSV & \textbf{0.62} & 0.66 & 0.69 \\
        \cmidrule(r){1-2} \cmidrule(l){3-5}
        \parbox[t]{2mm}{\multirow{6}{*}{\rotatebox[origin=c]{90}{Same Stance}}}
        & $\mathcal{S}$   & 0.85 & --    & *0.85 \\
        & $\mathcal{S}_0$ & 0.79 & *0.56 & *0.84 \\
        & $\mathcal{S}_D$ & 0.86 & --    & *0.86 \\
        & $\mathcal{P}$   & 0.86 & --    & *0.86 \\
        & $\mathcal{P}_0$ & 0.80 & 0.55  & *\textbf{0.88} \\
        & $\mathcal{P}_D$ & 0.86 & --    & *0.86 \\
        \bottomrule
    \end{tabular}
    }
    \caption{
    Evaluation of aggregation methods in ROC-AUC. 
    \textbf{Top}: evaluation on human annotation for \textit{Global} acceptability scores (\S\ref{sec:exp:partial_agreement}). \textbf{Middle}: evaluation on human annotation for \textit{Perspectivized} acceptability scores (\S\ref{sec:exp:aspect_of_disagreement}). \textbf{Bottom}: evaluation on \textit{Same Stance} prediction (\S\ref{sec:exp:noanno}). For fields marked with * lower acceptability scores mean the arguments are from the same stance.}

    \label{tab:acceptability_scores}
\end{table}

In our manual annotation, argument pairs of opposite stances had been annotated for global acceptability, i.e., being in i) partial agreement, ii) agreement, iii) disagreement or for being iv) orthogonal to each other. Since only two argument pairs were annotated with ``agreement'' (which is expected, since all annotated argument pairs are of opposite stance), we group ``agreement'' and ``partial agreement'' to form one agreement class. %

Tab.~\ref{tab:acceptability_scores} (top) shows the global performance of our methods. %
We report ROC-AUC, which allows us to compare our discrete and continuous aggregation methods using the same metric. 
We observe that detecting agreement is the most difficult: the best aggregation method ($\mathcal{S}_D$) achieves 0.60 AUC. For orthogonality and disagreement $\mathcal{P}_0$ performs best with 0.69 and 0.70 AUC, respectively. 

Fig.~\ref{app:fig:exp:psv_length} (\S\ref{app:sec:exp:psv_length}) shows scores for increasing lengths of PSVs, i.e., increasing number of perspectives the PSVs cover. %
For agreement and disagreement the performance is %
better for longer PSVs, while orthogonality is best with shorter PSVs. A closer look at the ROC curves (Fig.~\ref{app:fig:exp:ROC_curve}) reveals that shorter PSVs enable higher true positive rates for orthogonality. %
Fig.~\ref{app:fig:exp:psv_length} also shows the impact of filtering signature perspectives. %
Filtering by relevance shortens PSVs and improves agreement scores. %

\subsubsection{Perspectivized acceptability: Identifying aspects of (dis)agreement} \label{sec:exp:aspect_of_disagreement}
In \S\ref{sec:exp:partial_agreement} we predicted global agreement between arguments, classifying argument pairs as a whole. However, using PSVs, we can also identify which perspectives an argument pair agrees or disagrees on. Again, we compare to our human annotation. 

Tab.~\ref{tab:acceptability_scores} (middle) shows the results. %
$\mathcal{P}_0$ consistently performs  best for predicting agreeing, orthogonal and disagreeing perspectives, with 0.62, 0.75 and 0.76 AUC, respectively. Overall, the results tend to be better than global acceptability scores. This indicates that  averaging over all $n$ perspectives to obtain global scores incurs errors. A learned aggregation could alleviate this issue, but is beyond the scope of this work as it requires training data. %

\paragraph{Ablation without PSVs.} A valid concern regarding our approach is that our method reduces arguments to static vectors, which might oversimplify the nuances of deliberation. Further, it is of interest to what extent (dis)agreement scores could be predicted, in context, by a strong LLM. Thus, we also experimented with directly prompting \texttt{GPT4o} to predict the acceptability of two arguments on a specific perspective. This allows the model to directly compare the arguments, without the intermediate representation of PSVs. 
Note, however, 
that such an approach greatly diminishes \textit{interpretability}, given the lack of a structured representation and \textit{scalability}, as the number of comparisons scales quadratically with the number of arguments, as opposed to the linear scaling of our PSV framework. 

Our prompts include a short task description, the two arguments to be compared and the list of perspectives to be evaluated. The complete prompt and further details are in \S\ref{app:sec:ablation_pairwise_prompting}. 
\texttt{GPT4o} obtained 0.62, 0.66 and 0.69 ROC-AUC for agreement, orthogonality and disagreement, respectively (c.f.\ Table \ref{tab:acceptability_scores}). Surprisingly, this strong competitor falls behind 
our best-performing PSV aggregation $\mathcal{P}_0$. Prompting \texttt{GPT4o} with few-shot samples or chain-of-thought could potentially improve its results,
but will by no means justify the loss of interpretability and scalability that is inherent to our PSV approach. 

\begin{figure}
    \centering
     \includegraphics[width=1.0\linewidth]{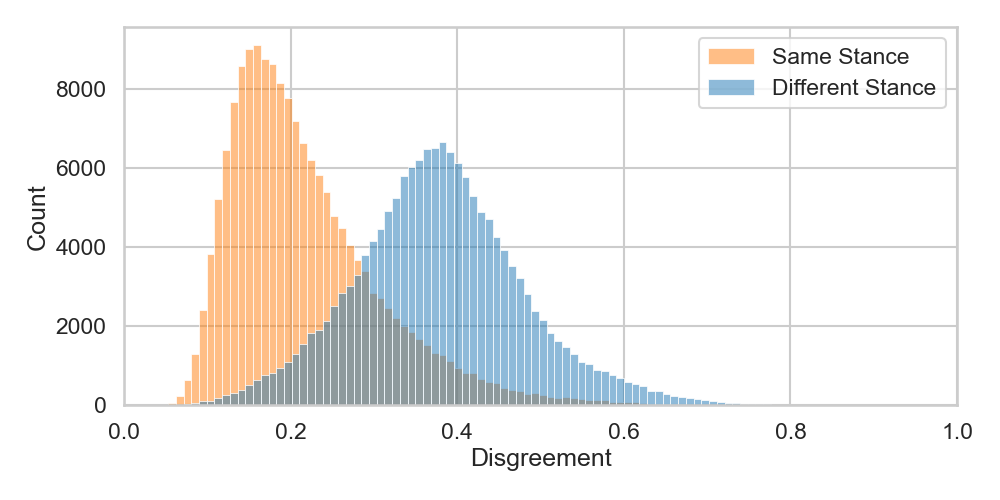}
    \caption{Disagreement ($\mathcal{P}_0$) distribution of argument pairs from the same stance or different stance.}
    \label{fig:histogram_same_stance}
\end{figure}

\subsubsection{Evaluation with unannotated data} %
\label{sec:exp:noanno}
So far we evaluated our methods on
our manually annotated data, which is naturally limited in size. To consolidate 
our analyses, we aim to 
verify our methods
on larger amounts of unannotated data.

\paragraph{Same stance.} To this end we perform
same-side classification: predicting whether two arguments from the same topic share the same stance.
We 
expect that arguments that share the same stance have higher agreement and lower disagreement scores, on average. Orthogonality is likely mostly independent of the stances from the argument pairs. 

Tab.~\ref{tab:acceptability_scores} (bottom) shows results for all arguments from the 
5 topics %
-- which amounts to %
326,836 argument pairs. %
As expected, we observe higher ROC-AUC values for predictions from agreement and disagreement scores 
compared to predictions from orthogonal scores. 
Also, disagreement using $\mathcal{P}_0$ performs best. This aligns with our previous results, where i) disagreement scores were higher than agreement scores and ii) $\mathcal{P}_0$ was the best aggregation method. Fig.~\ref{fig:histogram_same_stance} and \ref{app:fig:exp:same_stance} confirm that argument pairs of the same and diverging stances form distinct distributions for disagreement with $\mathcal{P}_0$.

\begin{figure}
    \centering
    \includegraphics[width=0.8\linewidth]{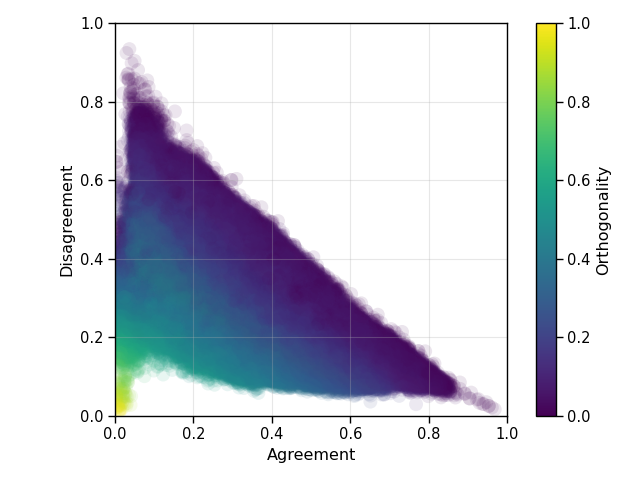}
    \caption{Acceptability scores computed with $\mathcal{P}_0$.}
    \label{fig:agreement_disagreement_scatter}
\end{figure}

\paragraph{Acceptability correlation.} As a final sanity check, in Fig.~\ref{fig:agreement_disagreement_scatter} we assess the correlation between acceptability scores by plotting the disagreement scores of argument pairs against their agreement scores. 
As expected, high agreement occurs with low disagreement and vice versa, while high orthogonality scores occur when both, agreement and disagreement, are low.

\begin{figure*}
    \centering    
     \includegraphics[width=0.8\linewidth]{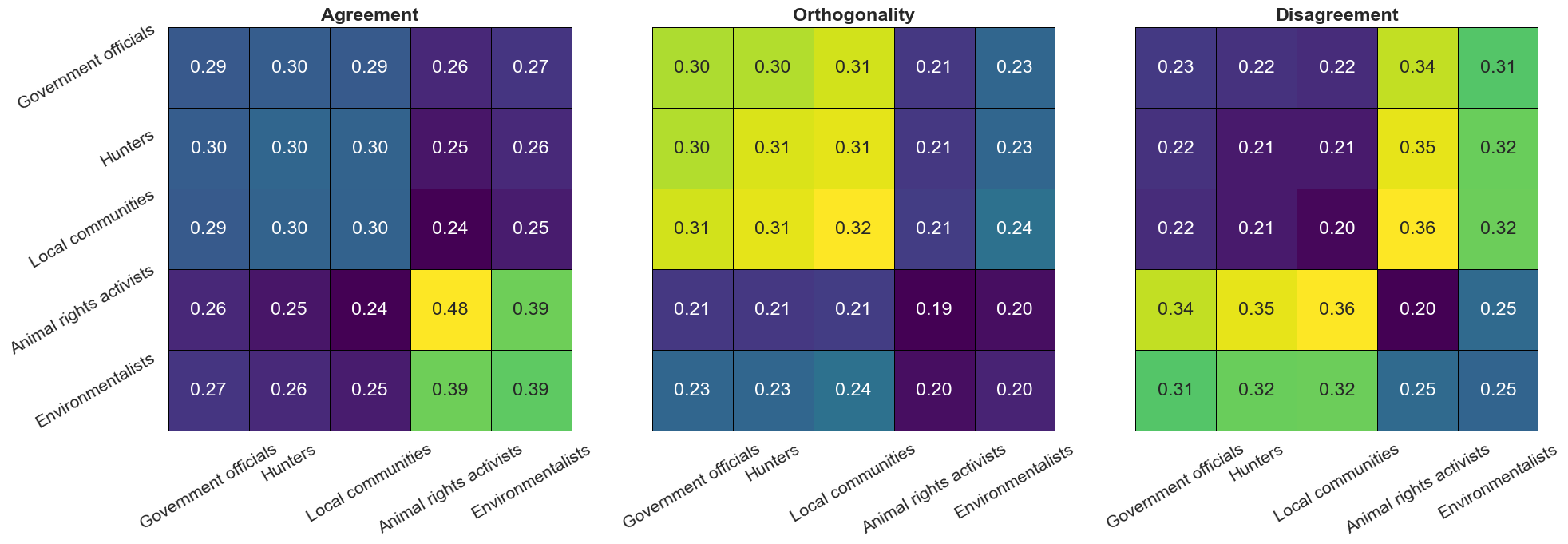}
    \caption{Agreement scores among stakeholder groups for `\textit{Animal Hunting}'. Fig.~\ref{app:fig:cs:stakeholder_global} shows results for other topics.}
    \label{fig:cs:stakeholder_global_AH}
\end{figure*}

\section{Case Study} \label{sec:case_study}
For our case study we construct PSVs with a signature of 100 perspectives and \texttt{GPT4o} (zero-shot) to predict PSV stance values. As aggregation method we use $\mathcal{P}_0$. 
We discuss our findings for the issue ``\textit{Should animal hunting be banned?}''. \S\ref{app:sec:case_study} shows results for all 5 topics from our annotation. 

We first look at global acceptability scores between different stakeholder groups, as shown in Figs.~\ref{fig:cs:stakeholder_global_AH} and \ref{app:fig:cs:same_stakeholder}. We observe that related stakeholder groups have higher agreement scores among each other, for example \textit{animal rights activists} and \textit{environmentalists}. Between %
these two and opposing stakeholder groups, such as \textit{hunters}, the disagreement scores are highest. Orthogonality scores are highest between the stakeholder groups \textit{government officials}, \textit{hunters} and \textit{local communities}. Indeed, \textit{government officials} and \textit{local communities} are vague and potentially diverse stakeholder groups for the given issue, which could explain why more arguments are orthogonal to each other. %

\begin{figure}
    \centering
    \includegraphics[width=0.8\linewidth]{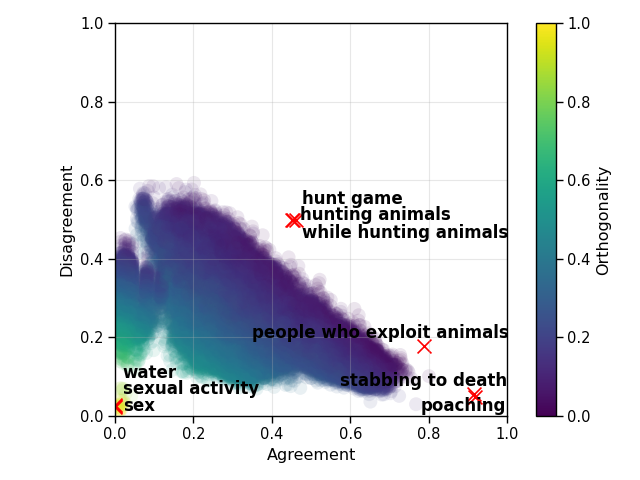}
    \caption{(Dis)agreement of selected perspectives.} 
    \label{fig:agreement_disagreement_scatter_perspectives_AnimalHunting}
\end{figure}

Fig.\ \ref{fig:agreement_disagreement_scatter_perspectives_AnimalHunting} and Tab.~\ref{app:tab:cs:concepts_by_stance} show the top-3 and top-5 perspectives per acceptability score (agreement, orthogonal, and disagreement). Across all arguments, the perspectives with highest agreement are perspectives that reflect a socially agreed `stigma',
such as \textit{poaching}, \textit{stabbing to death} or \textit{people who exploit animals} -- while perspectives with high disagreement reflect the overall stances in the debate (\textit{hunt game}, \textit{while hunting animals}). Orthogonal perspectives occur in only a few arguments, and hence are less relevant for the debated issue (\textit{sex}, \textit{sexual activity}, \textit{water}). 

Naturally, the agreement and disagreement perspectives depend on the overall stances of the compared arguments (cf.\ Tab.\ \ref{app:tab:cs:concepts_by_stance} and Fig.\ \ref{app:fig:cs:perspectives_by_same_stance}). When comparing arguments from opposing stances, the results remain in line with our findings in the previous paragraph. %
However, when the compared arguments have the same stance, the agreement and disagreement perspectives shift. For example \textit{hunt game}, which is a disagreement perspective across all arguments, is agreed upon among authors who are against banning animal hunting. They also agree on \textit{poaching}, showing that also people in favor of animal hunting disapprove poaching.\footnote{We identify that they are against poaching from the PSV stance values -- agreement scores alone do not express whether the agreement stems from positive or negative stance values.} %
Disagreement perspectives can reveal differences between arguers that share a global stance: \textit{control} and \textit{pleasure} are not agreed upon by all arguers %
in favor of hunting, showing that they are in favor of hunting for different reasons. %

Finally, we compare agreement and disagreement perspectives depending on the stakeholder groups we infer for the compared arguments. Here we can identify conflicts among (similar) stakeholder groups (for example \textit{killing for food} for \textit{animal rights activists} and \textit{environmentalists}) as well as common ground among (conflicting) stakeholder groups (for example \textit{poaching} or \textit{people who exploit animals} for \textit{hunters} and \textit{environmentalists}). 
Identifying such conflicts and shared understandings can help to better understand different opinions and hence, is a crucial step in deliberation. %

\section{Conclusions}

We present \textbf{P}erspectivized \textbf{S}tance \textbf{V}ectors (PSVs) -- a novel approach to represent fine-grained perspectives expressed in arguments on a debated topic. %
PSVs effectively identify and explain mutual %
(dis)\-agree\-ment between arguments and potential stakeholder groups, offering deep interpretability by revealing issue-specific perspectives driving such (dis)agreements. 
Identifying (dis)agreement perspectives can reveal the underlying reasons for conflicting viewpoints, 
and how they 
can potentially be resolved. 
Thus, we believe that our fine-grained analysis of perspectives using PSVs %
provides a valuable contribution to the growing field of deliberative decision making. %

\section*{Acknowledgments}
This work was funded by DFG, the German Research Foundation, for the project “ACCEPT: Perspectivized Argument Knowledge Graphs for Deliberation”, as part of the priority program ``RATIO: Robust Argumentation Machines'' (SPP-1999).

\section*{Limitations}
We evaluate our approach on \textsc{pakt}~\citep{plenz-etal-2024-pakt}, which is limited to English arguments from a predominantly US-context. As our PSV construction partially relies on LMs, it is to be expected that the quality of individual PSVs would be lower for data from a different background. However, our aggregation method is language- and culture agnostic and thus should be robust. 

Where possible we assess the quality of our approach automatically using data which is already available in large amounts (cf.\ \S \ref{sec:exp:noanno}). However, for fine-grained stance values and acceptability scores we had to rely on our manually annotated data. The annotated data covers 5 topics with 10 arguments each, which may seem like a rather small resource. However, collecting this data was a considerable annotation effort since we required annotations for argument pairs at the level of distinct perspectives. As our experiments are supported with a large-scale case study, we believe that our findings are reliable and trustworthy. 

Finally, predicting aspects of a debated issue which a group of arguments / authors agrees or disagrees on is a challenging task. Reducing this task to our PSV framework might cause oversimplifications. Nonetheless, we study these structured representations of arguments for two good reasons. First, they are highly interpretable -- which we believe to be important for deliberation tasks, to enhance (i) the trust of users and (ii) control for moderators. Second, large parts of our method have to be unsupervised due to a lack of training data. This makes training of end-to-end models infeasible. 
That being said, a more flexible end-to-end system might be able to obtain better performance in future work, for example, by creating larger amounts of partially annotated data, using our methods.

\bibliography{anthology,custom}

\FloatBarrier
\appendix

\section{Method}
\subsection{Signature} \label{app:sec:signature}
\paragraph{Concept selection} 
The commonsense knowledge graphs are taken from the published data of \citet{plenz-etal-2024-pakt}. Lemmatization was performed with the \texttt{en\_core\_web\_trf} model from Spacy. Future work could experiment with supervised concept selection, e.g., by finetuning models~\citep{plenz-frank-2024-graph} designed for knowledge graphs such as ConceptNet. 

\paragraph{Hypernym filtering.} We identify hypernyms using the NLTK implementation of WordNet. To allow for greater coverage we check for hypernyms within the lemmatized set of concepts. For each concept we only consider the first synset, and do not remove concepts which do not have a synset in WordNet. 

\paragraph{ChatGPT-based relevance filtering. } We use \texttt{ChatGPT-3.5-0125} to assign each concept with a score to reflect its \textit{relevance for a given issue}, 
using the following prompt:
\begin{quote}
    We plan to compare arguments depending on which concepts they evoke. Therefore, we created a catalog of concepts for each issue. For the following concept, decide whether it is relevant for the given issue: \\
    1: yes \\
    2: no \\
    Example Annotation for issue 'gun control': \\
    arm themselves:     1 \\
    control:            1 \\
    criminals:          1 \\
    dangerous:          1 \\
    laws regulate who:  1 \\
    own guns:           1 \\
    police:             1 \\
    politics:           1 \\
    shooting guns:      1 \\
    wrong:              1 \\
    \\
    Issue: \{\textit{debate topic}\} \\
    Concept: \{\textit{concept}\}
\end{quote}
The prompt is taken from our annotation guidelines. %
Depending on ChatGPT's output, we assign a binary label (\textit{relevant / irrelevant}) 
to each concept, for each issue.
The resulting labels are used as a filter to remove unrelated concepts.

To the best of our knowledge we are the first to use ChatGPT to assess the relevance of concepts for a given debate topic. %
Our human annotation indeed verifies that filtering with ChatGPT can boost precision for relevance (Table~\ref{tab:psv_signature}). 
In a more general scope, ChatGPT was successfully used for many argument classification tasks such as quality \citep{rocha-etal-2023-assessing,plenz-etal-2023-argument} and stance \citep{zhao-etal-2024-zerostance,zhang-etal-2024-how,zhang-etal-2023-investigating,plenz-etal-2023-argument} classification, which motivates our approach.

\subsection{Stance values} \label{app:sec:stance_values}

\paragraph{\texttt{RoBERTa}.}
We compose a synthetic dataset 
using the stance dataset of \citet{SobhaniMK16,MohammadSK17} and the dataset on human-values detection by \citet{mirzakhmedova-etal-2024-touche23-valueeval}. In the stance dataset, which is based on annotated tweets, we select those tweets that address the annotated target, by being against, in favor, or none of those %
(neutral). 
To increase the target diversity, we map each of the six targets to a hand-crafted set of synonyms and antonyms\footnote{In case of antonyms, we switch the classes against and in favor.}. To adapt the genre (from short tweets to more comprehensive arguments), we concatenate up to four tweets toward the same target to new instances\footnote{The aggregation of the stances is \textit{neutral} in case of a neutralized/balanced set of stances, \textit{in favor} if at least one stance is in favor and no stance is against, \textit{against} if at least one stance is against and no stance in favor, else the instance is dropped.}. We follow the same procedure with the human-values detection dataset on arguments, where we treat an annotated encouragement of a human value as \textit{in favor of} a set of hand-crafted associated concepts and \textit{against} a set of hand-crafted contrastive concepts. 
We consider a set of hand-crafted associated concepts of human values that are labeled as not relevant for this argument as the \textit{neutral} class. 

We start from a
\texttt{roberta-base} model that was already
fine-tuned on the task of sentiment prediction by \citet{loureiro-etal-2022-timelms}\footnote{The model is available from at \url{https://huggingface.co/cardiffnlp/twitter-roberta-base-sentiment-latest}. We also tried non-fine-tuned RoBERTa models, but they performed worse in preliminary analysis.}. We randomly split the synthetic dataset into 80\% training, 10\% development and 10\% test data, which we then use to further fine-tune the model with a learning rate of $2e-5$ and early stopping (evaluated after every 1000 processed train instances). We use a modified mean squared error as loss function:

\begin{equation}
    \begin{array}{rl}
         l= & \sum_{y_{\text{against}}}\lambda_{\text{against}} (\hat{y}_{\text{neutral}}^2 + \hat{y}_{\text{for}})^2\\
         + &  \sum_{y_{\text{neutral}}}\lambda_{\text{neutral}} (\hat{y}_{\text{against}} + \hat{y}_{\text{for}})^2\\
         + & \sum_{y_{\text{for}}}\lambda_{\text{for}} (\hat{y}_{\text{against}} + \hat{y}_{\text{neutral}}^2)^2 \\
    \end{array}
    \label{eq:FT-LM:loss}
\end{equation}

Here, $\lambda$ are hyperparameters to control the weighting between the three classes \textsc{Against}, \textsc{Neutral}, and \textsc{For}, where $y$ represents the target class and $\hat{y}_{c}$ the predicted probability ($[0, 1]$) for class $c$. 
We choose the model that performed best across 8 runs on the synthetic test split  
as the final model for \ac{PSV}-stance inference, once with $\lambda_{\text{against}}=\lambda_{\text{neutral}}=\lambda_{\text{for}}=1$, denoted as \emph{1-\texttt{RoBERTa}}, and once as an ensemble of three models which were trained with ($\lambda_{\text{against}}\in \{1, 1.33\}$, $\lambda_{\text{for}} \in \{1, 1.33\}$ and $\lambda_{\text{neutral}}=1$), denoted as \emph{3-\texttt{RoBERTa}}. 

\paragraph{\texttt{GPT4o}.}
We use \texttt{gpt-4o-2024-11-20} in a zero-shot setting (\emph{\texttt{GPT4o} (zero-shot)}) and a setting using 2 handcrafted examples showcasing that the overall sentiment of the argument does not need to correlate with the stance toward a specific concept (\emph{\texttt{GPT4o} (few-shot)}). The specific prompt used is: 

\begin{quote}
    Given a controversial topic discussed by a given argument, and an aspect, provide an one-word answer to the following question: Considering a person writing this argument, what is their attitude towards the given aspect: negative, neutral, or positive?
\end{quote}

\subsection{Aggregation Methods for \textit{Acceptability} Scores} \label{app:sec:method:aggregation_methods}
We first discuss the \textit{stance value-based} aggregation methods $\mathcal{S}$, $\mathcal{S}_0$ and $\mathcal{S}_D$. 

We use the  notation from  Tab.\  \ref{tab:aggregation_methods}, where  $s_i^j \in \{-1, 0, 1\}$ is the stance value of argument $j$ towards concept $i$.
Kronecker delta $\delta\mleft(x,y\mright)$ is 1 if $x=y$ and 0 else. 

\paragraph{Stance Value $\pmb{\mathcal{S}}$.} For agreement ($\mathcal{S}^+(s_i^1, s_i^2) = \delta\mleft(s_i^1, s_i^2\mright)$) we simply consider whether the arguments have the same stance value towards concept $c_i$ -- if yes, the agreement is 1 and otherwise 0. For disagreement ($\mathcal{S}^-(s_i^1, s_i^2) = 1 - \delta\mleft(s_i^1, s_i^2\mright)$) we instead check whether the stance values are different -- different values means the disagreement is 1, and otherwise it is 0. 

\paragraph{Stance Value (Considering Neutral) $\pmb{\mathcal{S}_0}$.} For this adaptation we consider the special role of neutral stance values: If both arguers are neutral towards a concept $c_i$, $c_i$ is not a meaningful indicator of agreement. Hence, we only record a perspectivised agreement %
if both stance values are in favor or against the concept -- but not if both are neutral: $\mathcal{S}_0^+(s_i^1, s_i^2) = \delta\mleft(s_i^1, s_i^2\mright)\left(1 - \delta\mleft(s_i^1, 0\mright)\right)$.

Similarly, we do not record disagreement
if at least one of the arguers is neutral towards a perspective $c_i$. Hence, we only 
record
disagreement if one stance value is in favor, while the other %
is against: $\mathcal{S}_0^-(s_i^1, s_i^2) = \left(1-\delta\mleft(s_i^1, s_i^2\mright)\right)\left(1 - \delta\mleft(s_i^1, 0\mright)\right)\left(1 - \delta\mleft(s_i^2, 0\mright)\right)$. 

For Orthogonality it is enough if at least one arguer is neutral towards the given perspective $c_i$. Hence, the orthogonality score is 1 if at least one arguer is neutral, and otherwise 0: $\mathcal{S}_0^\varnothing = \mathrm{min}\mleft(\sum\nolimits_{j=1}^2\delta\mleft(s^j_i, 0\mright), 1\mright)$

\paragraph{Stance Value (Difference) $\pmb{\mathcal{S}_D}$.} For agreement the previous approach might have the limitation that disagreement and orthogonal concepts both contribute 0 when computing a global agreement score by averaging over all concepts. It could be more expressive to have disagreement concepts \textit{reduce} the overall agreement, instead of treating them the same as orthogonal concepts. Hence, we designed $\mathcal{S}_D^+\mleft(s_i^1, s_i^2\mright) = \mathcal{S}_0^+\mleft(s_i^1, s_i^2\mright) - \mathcal{S}_0^-\mleft(s_i^1, s_i^2\mright)$ to be $+1$, $0$ and $-1$ for agreement, orthogonal and disagreement concepts, respectively. Analogously, we constructed $\mathcal{S}_D^-\mleft(s_i^1, s_i^2\mright) = \mathcal{S}_0^-\mleft(s_i^1, s_i^2\mright) - \mathcal{S}_0^+\mleft(s_i^1, s_i^2\mright)$. 

\paragraph{Stance Probability $\pmb{\mathcal{P}}$.} Aggregation methods that are based on \textit{stance probabilities}
have similar motivations. Their advantage, compared to the above methods based on stance values,
is that they return continuous values for each concept individually. 

For \textit{agreement}, we take the element-wise multiplication $\odot$ and sum the resulting values -- optionally disregarding
the neutrality scores. If the probability mass is similar (and not on neutral) for both arguments, then the agreement score is high. For \textit{orthogonality} we also consider element-wise multiplication. 

For \textit{disagreement}, we take the element-wise difference between the probabilities instead, and sum up the absolute values -- again, optionally without the neutral contribution. To obtain scores between 0 and 1 we normalize scores with a factor of $\nicefrac{1}{2}$.

\subsection{Stakeholder Annotation} \label{app:sec:stakeholder}

Debate portals are commonly lacking information about the authors of arguments, given privacy concerns.
To associate arguments with stakeholder information, 
we enrich our dataset by creating automatic predictions of stakeholder group information using ChatGPT, at issue and argument level..
To this end, we apply \texttt{ChatGPT-3.5-0125} for two subtasks:
First, we let ChatGPT predict 
potential stakeholder groups at \textit{issue-level}, by prompting ChatGPT to return a set of relevant stakeholder groups for a given topic:

\begin{quote}
    A stakeholder is a group of people who are affected by a topic. For example, the topic "Should young children have access to the internet?" has the stakeholders "Children" and "Parents". Return a list of the most important stakeholders for the topic "\{\textit{debate topic}\}". Return a simple list without explanations. Limit yourself to the few most important ones.
\end{quote}

We then use this set of stakeholder groups proposed by the model as input to a second call, 
to assign stakeholder types at \textit{argument-level}.
We ask ChatGPT to \textit{select}, from the given set of possible stakeholders, those types of stakeholders that could 
could plausibly utter a given argument from the relevant topic, using the following prompt:

\begin{quote}
    Here is an argument from someone: '\{\textit{argument}\}'. Which of these stakeholders are most likely to utter this argument: \{\textit{stakeholder set}\}? Return a list of stakeholders without additional information. Multiple may apply.
\end{quote}

We extract the stakeholder groups the model predicts to extend
each argument in our dataset. This typically results in 1--3 stakeholder labels per argument. 

As detailed below in \S\ref{app:sec:annotation}, we verify the predictions obtained by ChatGPT with manual annotations. Our evaluation shows that the predicted stakeholder groups are highly consistent with human judgment, at issue and argument level, and that only a small number of stakeholder groups is missing from ChatGPT's generated list.

\subsection{Ablation without PSV: Pairwise \texttt{GPT4o} prompting} \label{app:sec:ablation_pairwise_prompting}
We use \texttt{gpt-4o-2024-11-20}, which is the same model used to predict the perspectivized stances (c.f. \S\ref{sec:method:perspectivized_stances}). The prompt is 

\begin{quote}
    Arguments of opposite stance can have agreements – even though they don't agree on the issue at a binary level. Similarly, arguments with the same stance can disagree. We are interested in identifying and specifying such (dis-)agreements. We will present you with two independently written arguments of opposite stance and a list of concepts. 

        For each concept, annotate whether it is part of the agreement or disagreement: \\
        \textvisiblespace\textvisiblespace\textvisiblespace\textvisiblespace1: agreement, i.e., the authors could likely find agreement regarding this concept. \\
        \textvisiblespace\textvisiblespace\textvisiblespace\textvisiblespace2: neutral \\
        \textvisiblespace\textvisiblespace\textvisiblespace\textvisiblespace3: disagreement, i.e., it is not likely that the authors could agree regarding this concept. 

        Argument 1: \textit{\{argument\_1\}}

        Argument 2: \textit{\{argument\_1\}}

        Concepts: \textit{\{python\_list\_of\_concepts\}}

        Return your output as a list of integers, where each integer corresponds to the concept at the same index in the list of concepts. Do not include any additional information in your output. 
\end{quote}

\section{Experiments}
\subsection{Annotation} \label{app:sec:annotation}
Table~\ref{app:tab:annotation} shows the number of annotations and inter-annotator agreement (IAA) scores for different subtasks. IAA is measured using Krippendorff's $\alpha$~\citep{krippendorff-alpha}. 

We annotate data from 5 distinct topics:
\begin{enumerate}[label={\roman*)}, noitemsep, leftmargin=*]
    \item Should animal hunting be banned?
    \item Do you support the death penalty?
    \item Should students get paid for good grades?
    \item Should illegal immigrants be deported?
    \item Should kids have to wear school uniforms?
\end{enumerate}
The first topic was annotated by all 3 annotators, to assess IAA. The remaining topics were annotated by only one annotator, allowing us to collect more annotated data. All annotators are experts in computational linguistics and argumentation in particular. 

For each topic we annotate 30 signature concepts for i) their relevance with respect to the topic and ii) their granularity. These 30 concepts are the top-15 concepts per stance without any filtering. 
For each topic, we consider 10 arguments -- 5 from each stance. For each of these arguments, we annotate a topic-specific set of 10 concepts (top-5 concepts per stance, with hypernym filtering) for the \textit{PSV stance} (one of the values \textit{for, against, neutral}), yielding 500 annotations in total. 

To obtain annotations for \textit{agreement between arguments}, we annotate all arguments pairs of opposite stance within a topic, i.e., 25 argument-pairs per topic, on whether there is \textit{full} or \textit{partial agreement}, \textit{disagreement} or whether they are \textit{orthogonal} to each other. Further, for each of these argument-pairs, we annotate the same \textit{10 concepts} for whether the arguments (or rather the authors of the arguments) would \textit{agree, disagree} or are \textit{neutral} in relation to that concept. The annotated data as well as detailed annotation guidelines will be published upon acceptance. 

Table \ref{app:tab:annotation} summarizes the number of annotations and shows the IAA scores. We achieve moderate to substantial agreements, except for the concept-level argument pair annotation. 
We aimed for concise annotation guidelines to reduce the impact of subjective interpretations, but of course they can never be avoided in polarizing debates as we deal with. In particular for the concept-level argument pair annotation there are many subjective factors: interpretations of each of the arguments, as well as what their (dis)agreement is, and interpretation of the concept. This makes this a challenging annotation task, which partially explains the low IAA. However, we note that for disagreement the IAA is $\alpha=0.21$. Also, two annotators had an IAA of $\alpha=0.18$ across all classes. These annotators annotated 4 out of 5 topics. Future work can potentially further increase the IAA with multiple training rounds, thereby better calibrating annotators. 

\begin{table}
    \centering
    \resizebox{\linewidth}{!}{
    \begin{tabular}{lcc}
        \toprule
        Annotation task & \# annotations & IAA $\alpha$\\
        \cmidrule(r){1-1} \cmidrule(l){2-3}
        Signature -- relevance & \phantom{1,}150 & 0.42 \\
        Signature -- granularity & \phantom{1,}150 & 0.42 \\
        PSV stances & \phantom{1,}500 & 0.60 \\
        Argument pairs & \phantom{1,}125 & 0.64 \\
        Argument pairs -- Concept-level & 1,250 & 0.03 \\
        \bottomrule
    \end{tabular}
    }
    \caption{Number of annotations and the inter-annotator agreement (IAA) measured in Krippendorff's $\alpha$ for different annotation tasks. Note that Krippendorff's $\alpha$ is computed only on one topic, i.e., only on one fifth of the total number of annotations shown in this table. }
    \label{app:tab:annotation}
\end{table}

We also obtain validating annotations for the stakeholder group predictions that ChatGPT generated at  issue- and argument level, (see  \S\ref{app:sec:stakeholder}).

The stakeholder groups that the model generates as potentially relevant for an issue were judged for \textit{relevancy} at issue level.
At argument level, the applicability of a stakeholder
to a given argument can be labeled as  \textit{plausible}, \textit{unlikely} or as \textit{independent} (meaning that there is no consensus in relating  the argument's content to a stakeholder group) -- depending on how likely it is that members of a given group are to utter it. The distribution of these %
labels 
is displayed in Table \ref{app:tab:stakeholder-frequency}.

\begin{table}[]
    \centering
     \resizebox{0.6\linewidth}{!}{
    \begin{tabular}{lc}
    \toprule
        Label & Frequency [\%] \\ \cmidrule(r){1-1} \cmidrule(l){2-2}
        Plausible & 29.5 \\ %
        Independent & 46.4 \\ %
        Unlikely & 24.1 \\ %
    \bottomrule
    \end{tabular}}
    \caption{Distribution of stakeholder group labels at argument-level(ground truth via majority voting). 
}
    \label{app:tab:stakeholder-frequency}
\end{table}

\begin{table}[]
    \centering
         \resizebox{0.53\linewidth}{!}{
    \begin{tabular}{ccc}
        \toprule
        Precision & Recall & F1 \\ \midrule
        1.0 & 78.6 & 88.0 \\
        \bottomrule
    \end{tabular}}
    \caption{Evaluation of ChatGPT-generated stakeholder annotations on issue-level.}
    \label{app:tab:topicstakeholder}
\end{table}

For the five \textit{issues} included in
our annotation, ChatGPT generated on average five possibly relevant stakeholder groups. Our annotators were asked to
assign a relevancy label for each generated stakeholder group for a given issue. We also asked them
to list any missing stakeholders that could be relevant to the topic. These are interpreted as missing, allowing us to assess recall. %
Results are displayed in Table \ref{app:tab:topicstakeholder}. While a small amount of relevant stakeholders are missing from ChatGPT's output, all predicted stakeholder groups were validated
as being relevant to the larger topic by the majority of annotators. 

The validation of predicted stakeholder groups at \textit{argument-level} was performed by human labelers.   
If annotators disagree on a 
label, the gold label is determined via majority voting. We compare this ground truth to the predictions of ChatGPT to determine whether its predictions of plausible stakeholder groups at argument level are  
valid or invalid.

Note, however, that ChatGPT's predictions were restricted to the binary classes \textit{plausible} and \textit{unlikely}, while the annotators were offered an additional label to express that an argument may be \textit{independent} from a given stake holder group (see Table \ref{app:tab:stakeholder}).
Most indicative of the quality of ChatGPT’s prediction is thus Recall for the relevant classes \textit{unlikely} and \textit{plausible}, which show substantial and moderate IAA (Fleiss’ Kappa
between three annotators), respectively, as opposed to \textit{independent} which achieves only fair IAA quality.

\begin{table}[]
    \centering
    \resizebox{0.85\linewidth}{!}{
    \begin{tabular}{lccccc}
        \toprule
        Class & Prec. & Rec. & F1 & IAA \\ \cmidrule(r){1-1} \cmidrule(l){2-5}
        Plausible & 38.7 & 63.1 & 48.0 & 0.447 \\
        Independent & \phantom{0}0.0 & \phantom{0}0.0 & \phantom{0}0.0 & 0.228 \\
        Unlikely & 31.6 & 67.9 & 43.1 & 0.648 \\
        \bottomrule
    \end{tabular}
    }
    \caption{Evaluation of ChatGPT's stakeholder groups prediction for arguments  against human annotation. While ChatGPT was asked for binary judgements  (\textit{plausible, unlikely}), human annotators were also offered a third category \textit{independent}. Most indicative is ChatGPT's Recall for the relevant classes \textit{unlikely} and \textit{plausible}, with substantial and moderate IAA (Fleiss' Kappa between three annotators), as opposed to \textit{independent}.   
    }
    \label{app:tab:stakeholder}
\end{table}

\subsection{Confusion matrices for PSV stance value prediction} \label{app:sec:exp:stance_value}
\begin{figure}
    \centering
     
     \includegraphics[width=\linewidth]{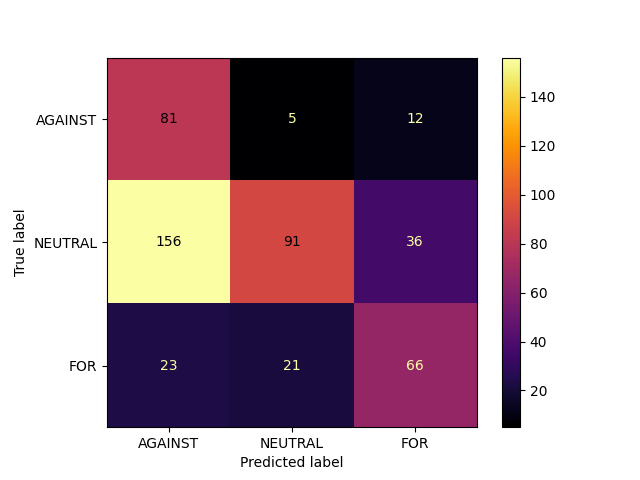}
    \caption{\texttt{GPT4o} (zero-shot) confusion matrix for stance value prediction compared to gold annotation.}
    \label{app:fig:confusion_matrix_stance_value_gpt4o}
\end{figure}

Figure \ref{app:fig:confusion_matrix_stance_value_gpt4o} shows the confusion matrix for \texttt{GPT4o} (zero shot). 

\subsection{Impact of PSV length on Acceptability scores} 
\begin{figure}
    \centering
    \includegraphics[width=\linewidth]{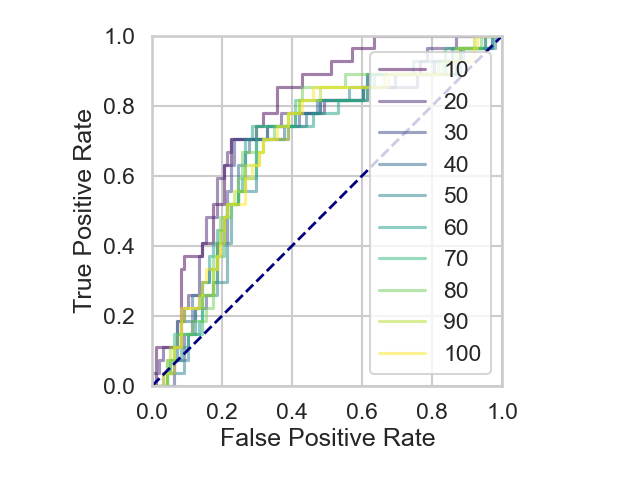}
    \caption{ROC curves for Orthogonality with $\mathcal{P}_0$ for different PSV lengths. }
    \label{app:fig:exp:ROC_curve}
\end{figure}

\begin{figure*}
    \centering
     \begin{subfigure}[b]{0.8\textwidth}
         \centering
         \includegraphics[width=\textwidth]{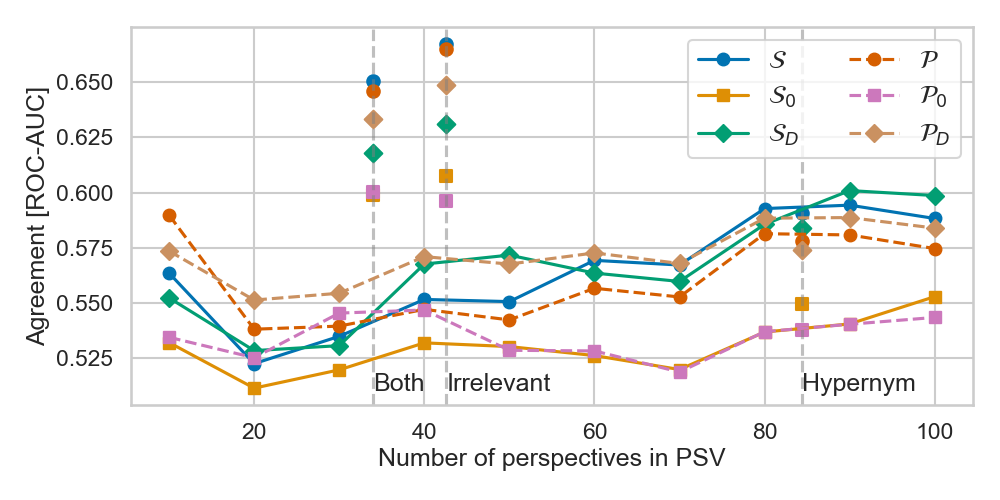}
         \caption{Agreement.}
     \end{subfigure}
     \\
     \begin{subfigure}[b]{0.8\textwidth}
         \centering
         \includegraphics[width=\textwidth]{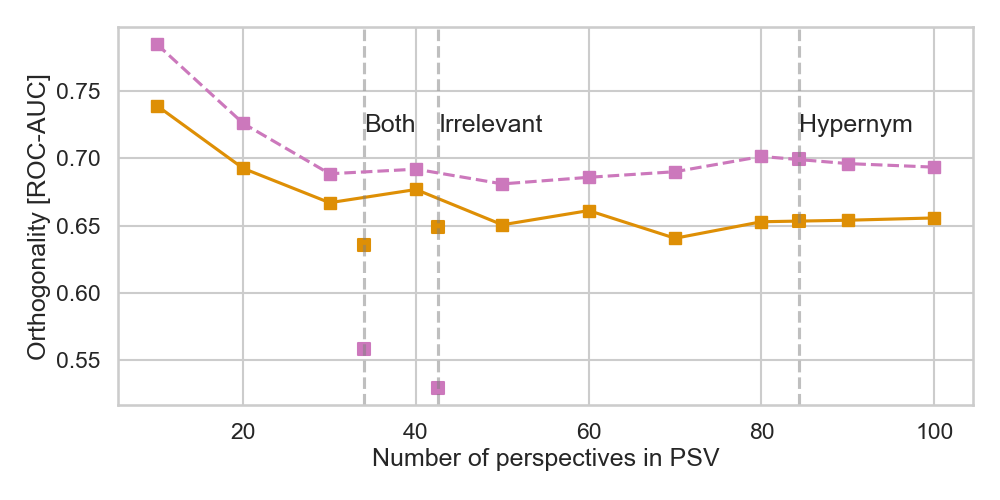}
         \caption{Orthogonality.}
     \end{subfigure}
     \\
     \begin{subfigure}[b]{0.8\textwidth}
         \centering
         \includegraphics[width=\textwidth]{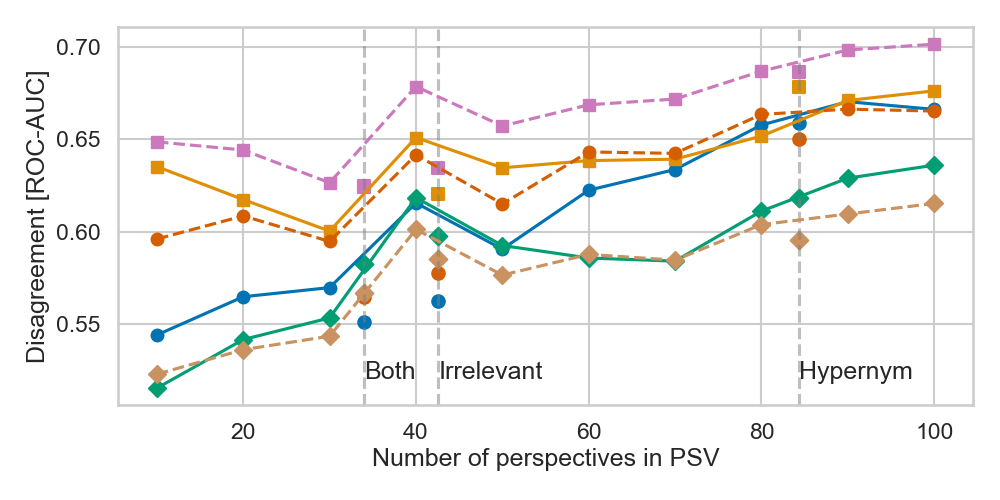}
         \caption{Disagreement.}
     \end{subfigure}
    \caption{ROC-AUC scores compared to human annotation depending on PSV length. Filtering options are shown at their average PSV length across the 5 topics. }
    \label{app:fig:exp:psv_length}
\end{figure*}

Figure~\ref{app:fig:exp:psv_length} shows the impact of the number of perspectives (i.e., the dimension or length of a PSV) on global agreement, orthogonality and disagreement prediction. Figure~\ref{app:fig:exp:ROC_curve} shows corresponding ROC curves for orthogonality. 

\label{app:sec:exp:psv_length}

\subsection{Same stance prediction} \label{app:sec:same_stance}
\begin{figure*}
    \centering
     \begin{subfigure}[b]{0.49\textwidth}
         \centering
         \includegraphics[width=\textwidth]{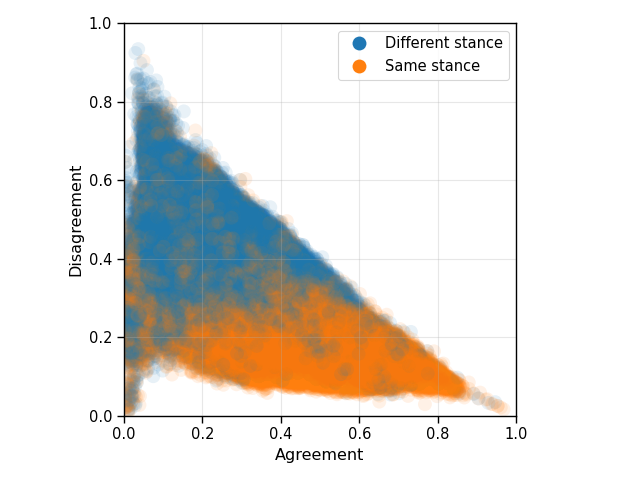}
         \caption{All topics.}
     \end{subfigure}
     \hfill
     \begin{subfigure}[b]{0.49\textwidth}
         \centering
         \includegraphics[width=\textwidth]{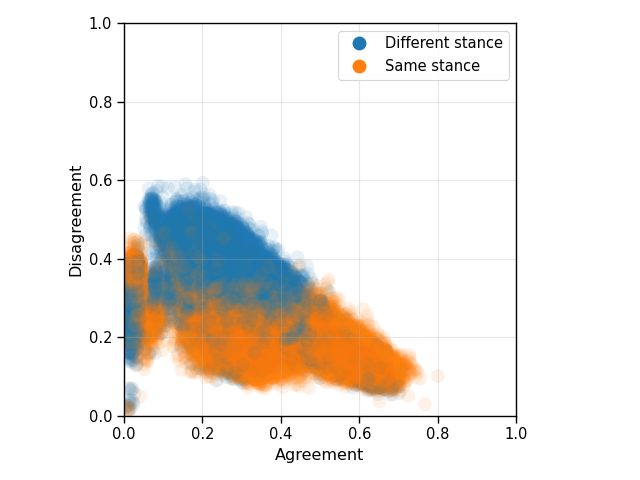}
         \caption{Animal Hunting.}
     \end{subfigure}
     \\
     \begin{subfigure}[b]{0.49\textwidth}
         \centering
         \includegraphics[width=\textwidth]{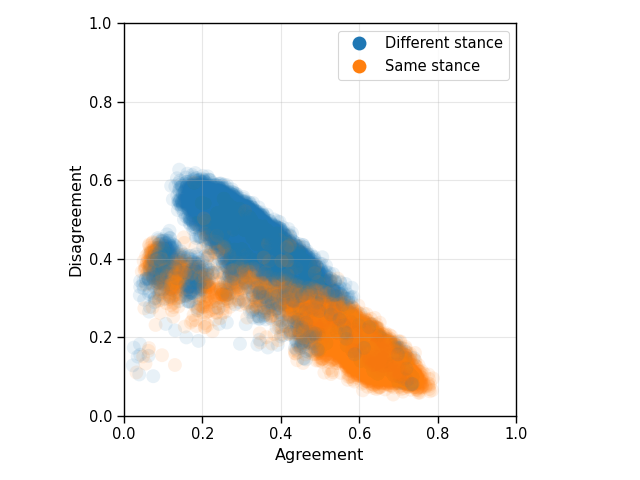}
         \caption{Death Penalty.}
     \end{subfigure}
     \hfill
     \begin{subfigure}[b]{0.49\textwidth}
         \centering
         \includegraphics[width=\textwidth]{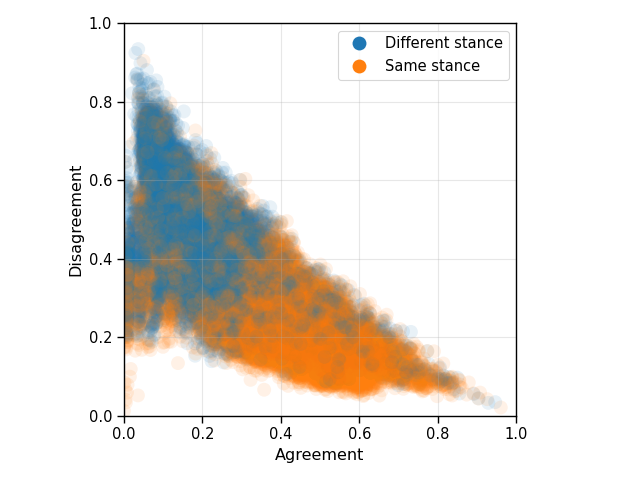}
         \caption{Good Grades.}
     \end{subfigure}
     \\
     \begin{subfigure}[b]{0.49\textwidth}
         \centering
         \includegraphics[width=\textwidth]{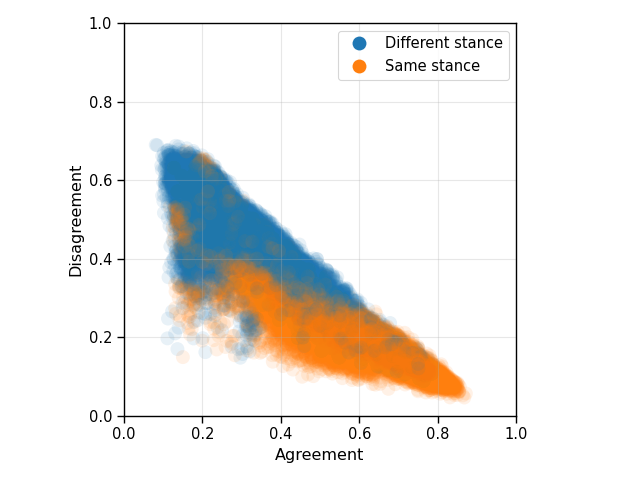}
         \caption{Illegal Immigrants.}
     \end{subfigure}
     \hfill
     \begin{subfigure}[b]{0.49\textwidth}
         \centering
         \includegraphics[width=\textwidth]{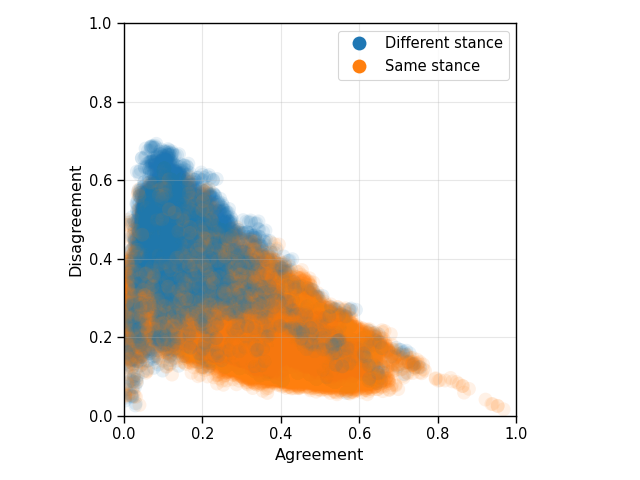}
         \caption{School Uniforms.}
     \end{subfigure}
    \caption{Argument pairs depending on their (Dis)agreement scores, colored by same stance.}
    \label{app:fig:exp:same_stance}
\end{figure*}

Figure \ref{app:fig:exp:same_stance} shows the (dis)agreement distributions for argument pairs of same and different stances. 

\section{Case Study} \label{app:sec:case_study}
\begin{figure*}
    \centering
     \begin{subfigure}[b]{0.7\textwidth}
         \centering
         \includegraphics[width=\textwidth]{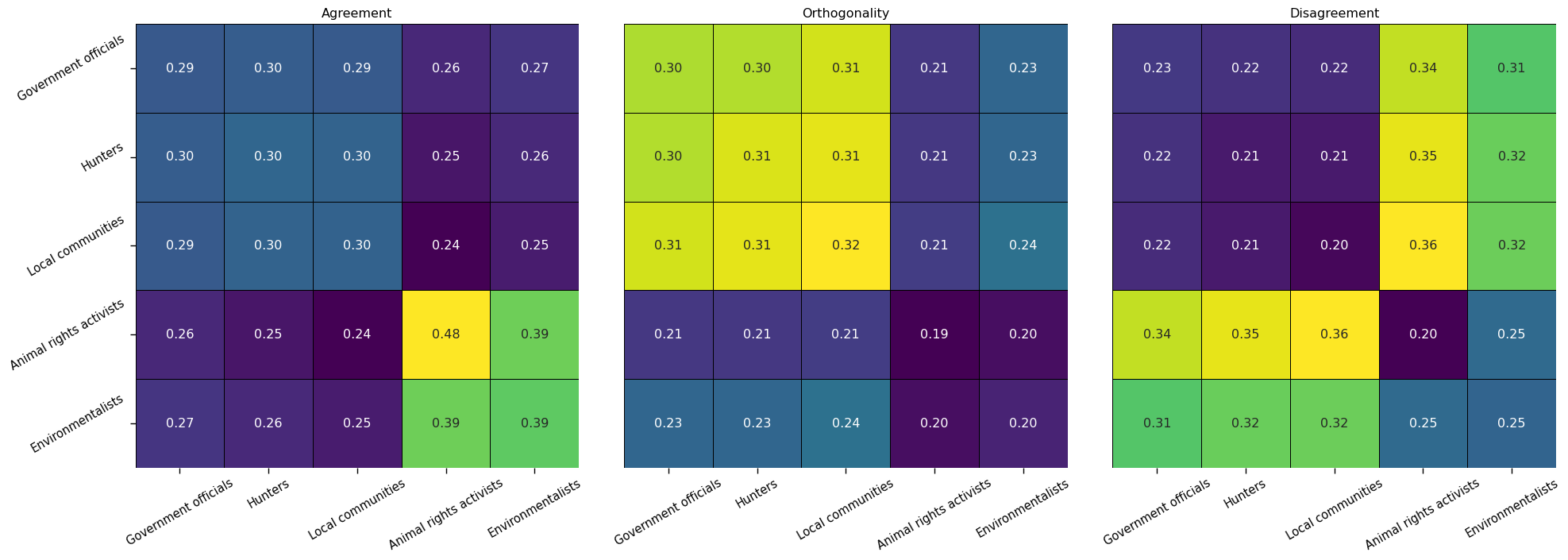}
         \caption{Animal Hunting.}
     \end{subfigure}
     \\
     \begin{subfigure}[b]{0.8\textwidth}
         \centering
         \includegraphics[width=\textwidth]{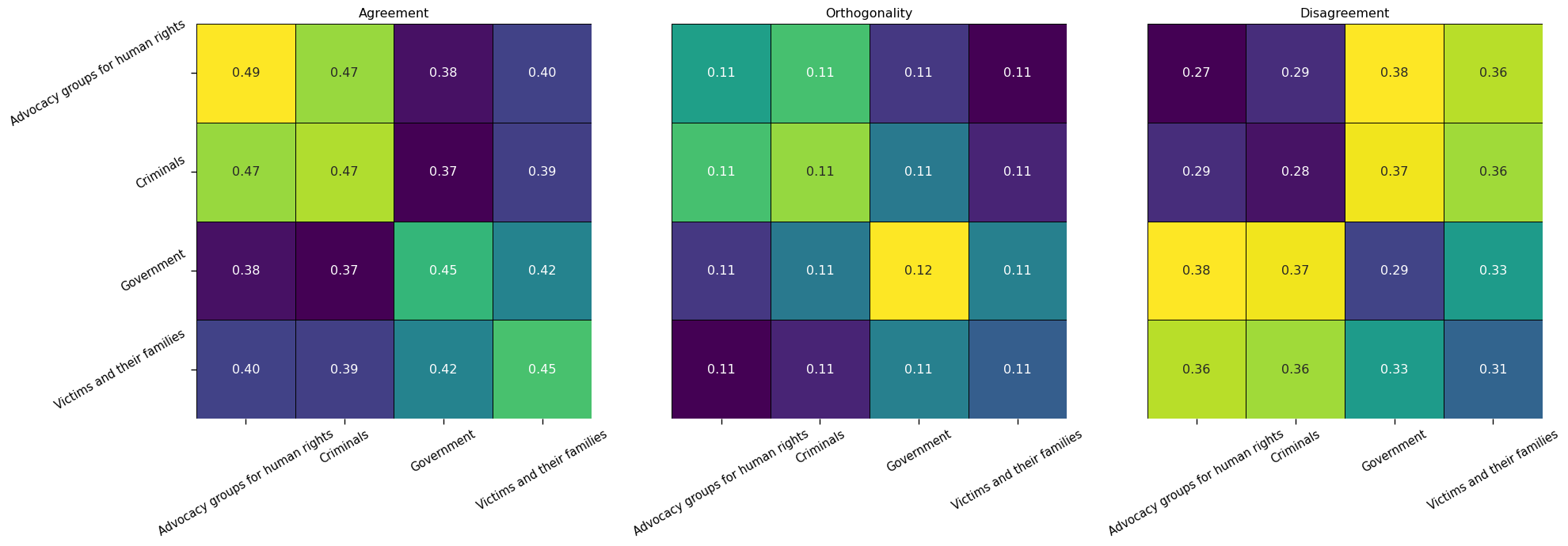}
         \caption{Death Penalty.}
     \end{subfigure}
     \\
     \begin{subfigure}[b]{0.7\textwidth}
         \centering
         \includegraphics[width=\textwidth]{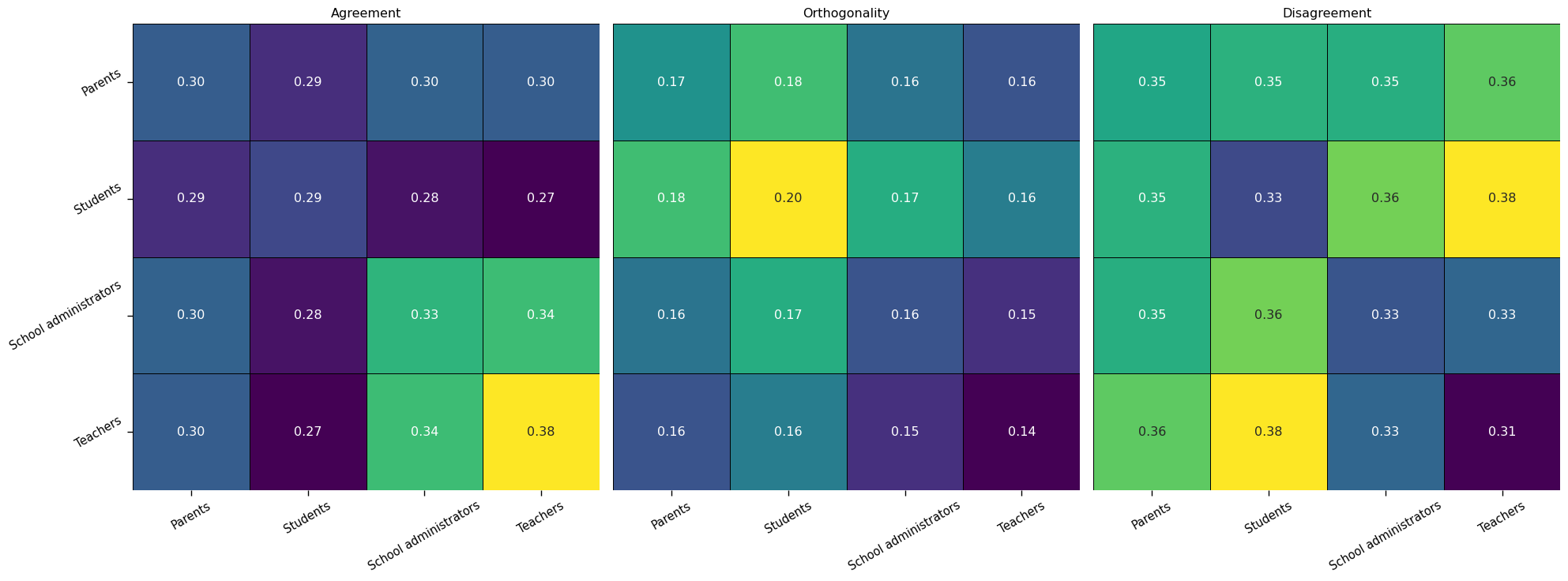}
         \caption{Good Grades.}
     \end{subfigure}
     \\
     \begin{subfigure}[b]{0.7\textwidth}
         \centering
         \includegraphics[width=\textwidth]{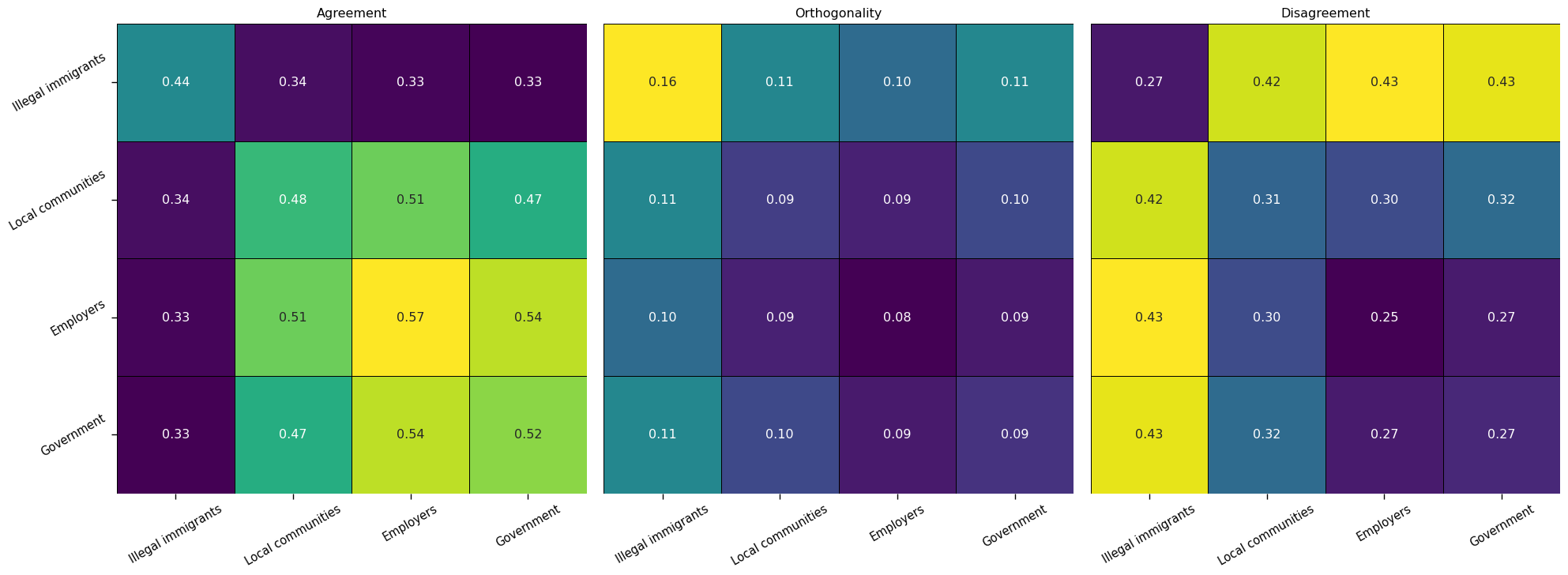}
         \caption{Illegal Immigrants.}
     \end{subfigure}
     \\
     \begin{subfigure}[b]{0.7\textwidth}
         \centering
         \includegraphics[width=\textwidth]{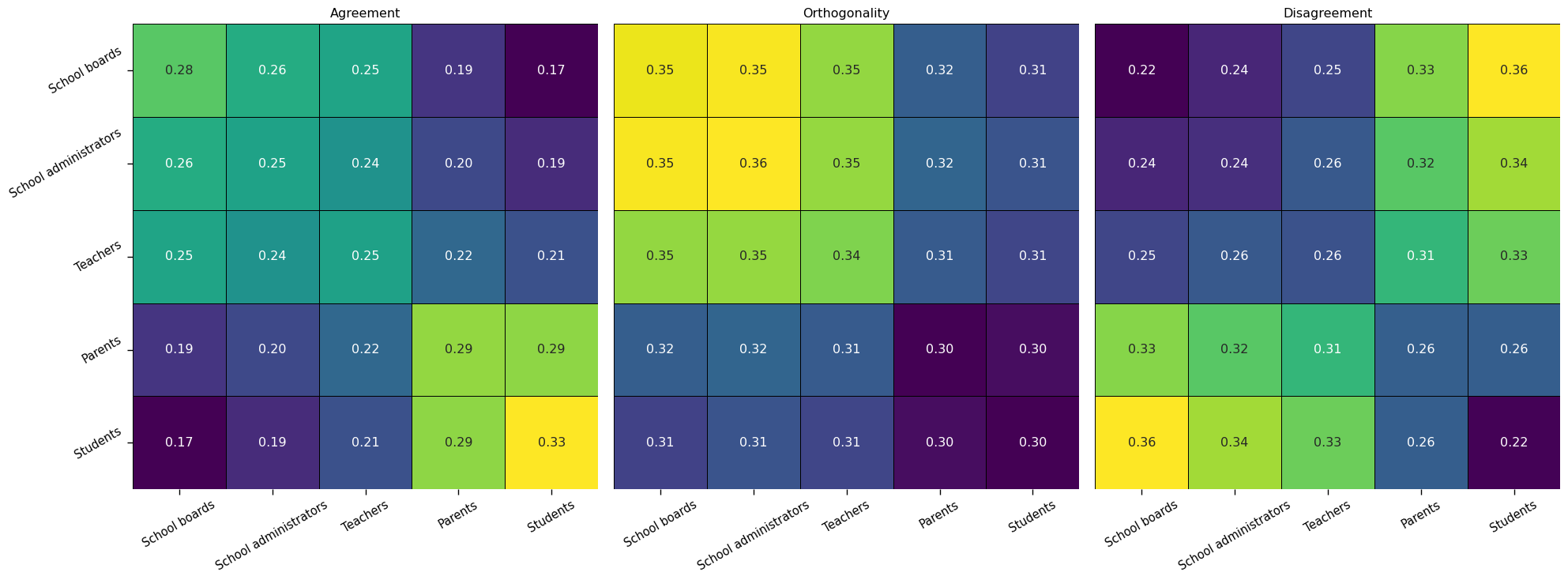}
         \caption{School Uniforms.}
     \end{subfigure}
    \caption{Acceptability scores among stakeholder groups for different topics.}
    \label{app:fig:cs:stakeholder_global}
\end{figure*}

\begin{figure*}
    \centering
     \begin{subfigure}[b]{0.49\textwidth}
         \centering
         \includegraphics[width=\textwidth]{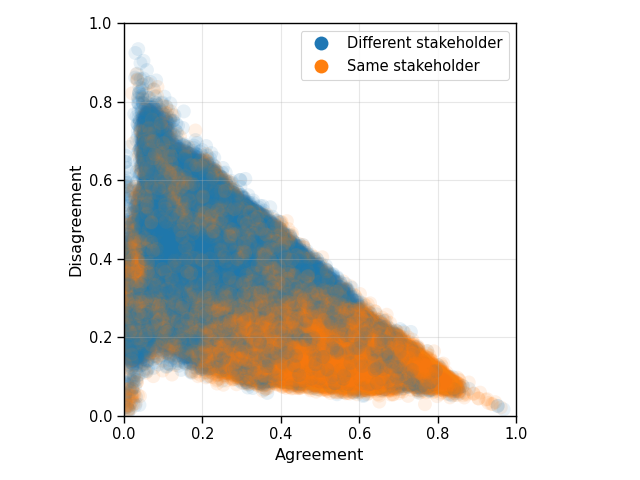}
         \caption{All topics.}
     \end{subfigure}
     \hfill
     \begin{subfigure}[b]{0.49\textwidth}
         \centering
         \includegraphics[width=\textwidth]{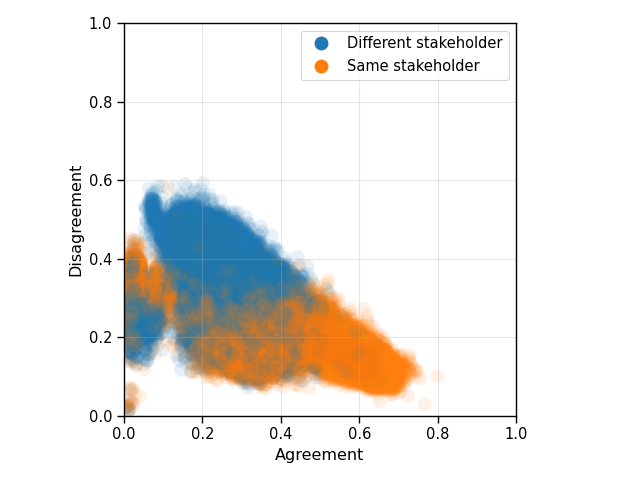}
         \caption{Animal Hunting.}
     \end{subfigure}
     \\
     \begin{subfigure}[b]{0.49\textwidth}
         \centering
         \includegraphics[width=\textwidth]{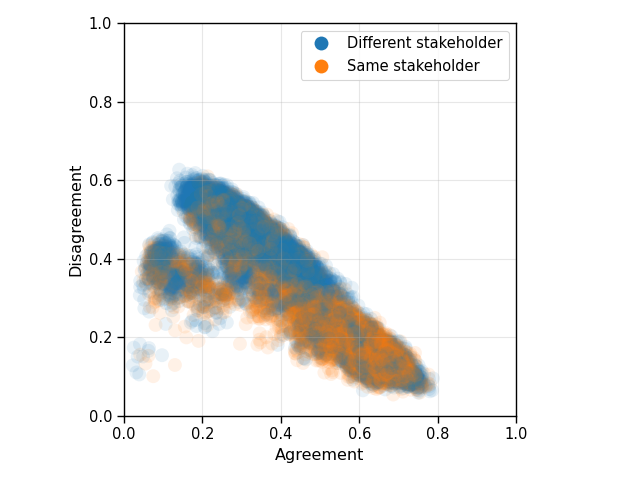}
         \caption{Death Penalty.}
     \end{subfigure}
     \hfill
     \begin{subfigure}[b]{0.49\textwidth}
         \centering
         \includegraphics[width=\textwidth]{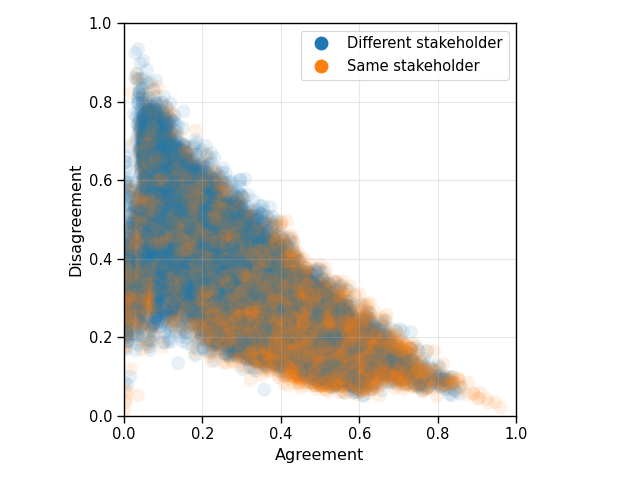}
         \caption{Good Grades.}
     \end{subfigure}
     \\
     \begin{subfigure}[b]{0.49\textwidth}
         \centering
         \includegraphics[width=\textwidth]{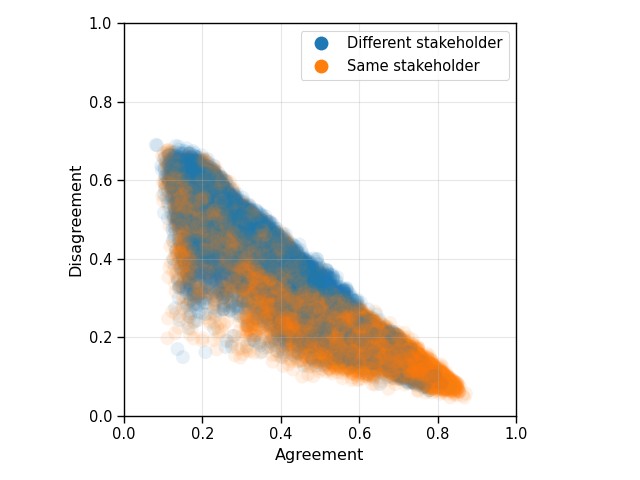}
         \caption{Illegal Immigrants.}
     \end{subfigure}
     \hfill
     \begin{subfigure}[b]{0.49\textwidth}
         \centering
         \includegraphics[width=\textwidth]{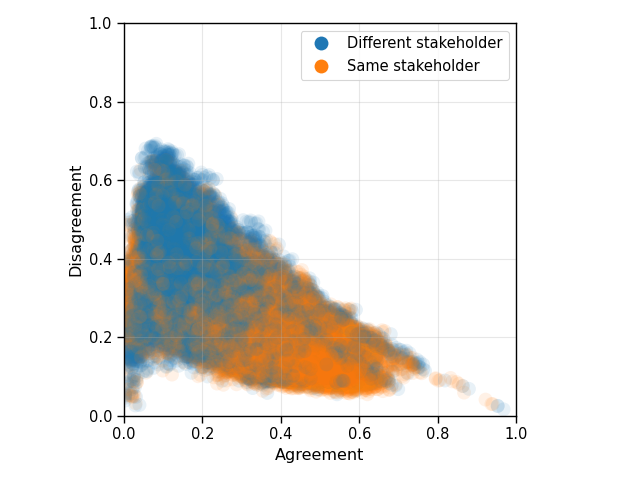}
         \caption{School Uniforms.}
     \end{subfigure}
    \caption{Argument pairs depending on their (Dis)agreement scores, colored by whether the stakeholders of the two arguments are the same. As our stakeholder prediction returns a set of stakeholders for each argument, we check whether one set is a subset of the other. }
    \label{app:fig:cs:same_stakeholder}
\end{figure*}

\begin{figure*}
    \centering
     \begin{subfigure}[b]{0.49\textwidth}
         \centering
         \includegraphics[width=\textwidth]{imgs/agreement_disagreement_orthogonality_scatter_perspectives_AnimalHunting.png}
         \caption{All argument pairs.}
     \end{subfigure}
     \hfill
     \begin{subfigure}[b]{0.49\textwidth}
         \centering
         \includegraphics[width=\textwidth]{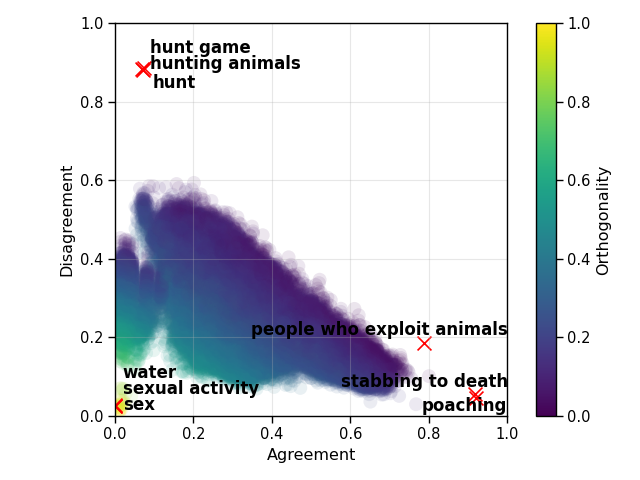}
         \caption{Opposing Stance.}
     \end{subfigure}
     \\
     \begin{subfigure}[b]{0.49\textwidth}
         \centering
         \includegraphics[width=\textwidth]{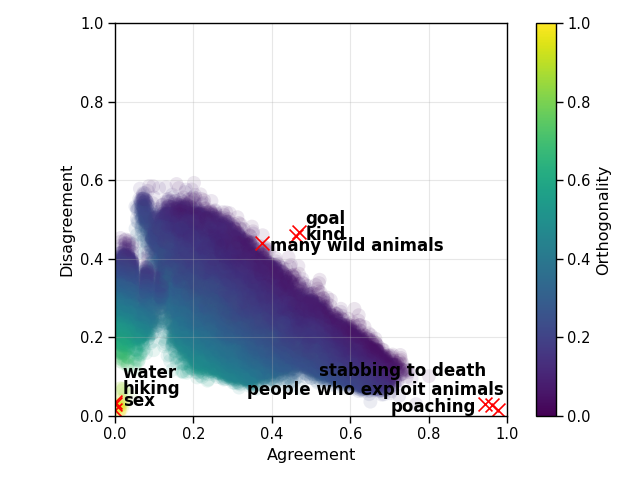}
         \caption{Same Stance: \textit{yes}.}
     \end{subfigure}
     \hfill
     \begin{subfigure}[b]{0.49\textwidth}
         \centering
         \includegraphics[width=\textwidth]{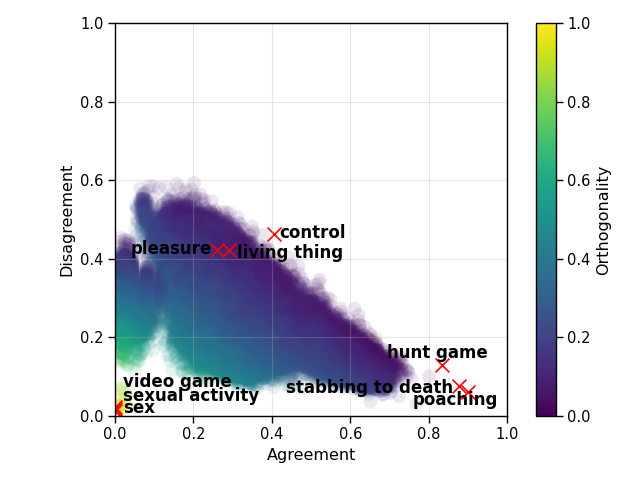}
         \caption{Same Stance: \textit{no}.}
     \end{subfigure}

    \caption{Most prominent perspectives for Animal Hunting, depending on the stances of the compared arguments.}
    \label{app:fig:cs:perspectives_by_same_stance}
\end{figure*}

This section collects plots and tables for the case study. 
Figure \ref{app:fig:cs:stakeholder_global} presents acceptability scores between different stakeholders. 
Figure \ref{app:fig:cs:same_stakeholder} shows argument pairs depending on whether the respective stakeholders are the same or different. 
Table \ref{app:tab:cs:concepts_by_stance} and Figure \ref{app:fig:cs:perspectives_by_same_stance} show the perspectives with highest acceptability scores depending on the topic and argument stances. 
Similarly, Tables \ref{app:tab:cs:concepts_by_stakeholder_AnimalHunting}-\ref{app:tab:cs:concepts_by_stakeholder_SchoolUniforms} show the most prominent perspectives depending on the stakeholder groups of the arguments. 

\begin{table*}
\centering
\resizebox*{!}{0.95\textheight}{
\begin{tabular}{p{0.2cm}p{2cm}|p{3cm}p{3cm}p{3cm}p{3cm}}
\toprule
    & & all & both pro & both con & one pro one con \\ \midrule
        \parbox[t]{2mm}{\multirow{13}{*}{\rotatebox[origin=c]{90}{Animal Hunting}}}
     & Agreement & poaching, stabbing to death, people who exploit animals, evil, unnatural thing & people who exploit animals, stabbing to death, poaching, blood sport, cruelty & poaching, stabbing to death, hunt game, hunt, wrong & poaching, stabbing to death, people who exploit animals, unnatural thing, evil \\
     & Orthogonality & sex, sexual activity, water, hiking, video game & water, hiking, sex, sexual activity, city & sex, sexual activity, video game, copulating, fly & sex, water, sexual activity, hiking, video game \\
     & Disagreement & hunt game, while hunting animals, hunting animals, hunter, hunt & goal, kind, many wild animals, endangered species, animals and sometimes people & control, pleasure, living thing, kind, humans & hunt game, hunting animals, hunt, hunter, while hunting animals \\
\midrule 
    \parbox[t]{2mm}{\multirow{13}{*}{\rotatebox[origin=c]{90}{Death Penalty}}}
     & Agreement & committing crime, stupid, murdering, murder, injustice & someones who commits murder, committing crime, human right, murderers, crimes & murdering, stupid, kill, human killing, injustice & committing crime, stupid, murdering, murder, crimes \\
     & Orthogonality & album, legs, british, running after ball, play golf & british, album, legs, running after ball, play golf & album, play golf, legs, running after ball, british & album, legs, british, running after ball, play golf \\
     & Disagreement & capital punishment, death sentence, death penalty, right to life, face death penalty & legal, law, change, killed, kill & innocent people, change, human, imprisonment, prosecuted and sent to jail & death sentence, sentenced to death, death penalty, face death penalty, capital punishment \\
\midrule 
    \parbox[t]{2mm}{\multirow{11}{*}{\rotatebox[origin=c]{90}{Good Grades}}}
     & Agreement & pay off teacher, bank on failing in school, bribe, penalty, free & bank on failing in school, pay off teacher, reward, get paid, money & bribe, pay off teacher, twenty bucks for every, buying, get paid & pay off teacher, bank on failing in school, bribe, penalty, free \\
     & Orthogonality & sleep, sports, eat, clean house, acting in play & sports, clean house, eat, acting in play, sleep & sleep, sports, eat, clean house, acting in play & sports, sleep, eat, clean house, acting in play \\
     & Disagreement & reward, make money, get paid, value, money & satisfaction, school, fee, learning, discipline & education, further education, learn lessons well, educate, study & get paid, money, reward, make money, twenty bucks for every \\
\midrule 
    \parbox[t]{2mm}{\multirow{11}{*}{\rotatebox[origin=c]{90}{Illegal Immigrants}}}
     & Agreement & unnatural thing, stealing, criminals, criminal act, people who break laws & people who break laws, illegal, stealing, amnesty, exemption & turn away, deportations, deport, ejection, go home & unnatural thing, stealing, criminals, government, criminal act \\
     & Orthogonality & video game, computing, canada, food, walking & video game, computing, canada, food, walking & computing, video game, walking, canada, food & video game, computing, canada, food, walking \\
     & Disagreement & turn away, ouster, exile, order, return home & law, immigration law, exile, country, order & quality, america, good feelings, exemption, change of location & deportations, deport, amnesty, immigrants, immigrants people who \\
\midrule 
    \parbox[t]{2mm}{\multirow{14}{*}{\rotatebox[origin=c]{90}{School Uniforms}}}
     & Agreement & bad, disguise, touchy about wearing uniforms, pain, special outfit & uniform, school uniform, bad, reason, required for schools to function effectively & school uniform, uniform, touchy about wearing uniforms, required for schools to function effectively, disguise & bad, disguise, very expensive, kids clothing, change \\
     & Orthogonality & food, mathematics, dance, painting, church & church, painting, food, fencing, biology & mathematics, food, dance, sports, plastic surgery & food, dance, mathematics, church, painting \\
     & Disagreement & motivation, expressing yourself, reason, ideal, self esteem & special way to dress, kids clothing, clothes, clothing, changing appearance & change, fashion, expressing yourself, student, dress themselves & school uniform, uniform, required for schools to function effectively, school, ideal \\

\bottomrule
\end{tabular}
}
\caption{Concepts with highest acceptability scores depending on the topic and argument stances.}
\label{app:tab:cs:concepts_by_stance}
\end{table*}

\begin{table*}
\centering
\resizebox*{\linewidth}{!}{
\begin{tabular}{p{3cm}|p{3cm}p{3cm}p{3cm}p{3cm}p{3cm}}
\toprule
    & Government officials & Hunters & Local communities & Animal rights activists & Environmentalists \\ \midrule
    Government officials & poaching, stabbing to death, wrong, hunt game, evil & poaching, stabbing to death, hunt game, hunt, hunting animals & poaching, stabbing to death, hunt game, hunt, hunting animals & stabbing to death, poaching, evil, unnatural thing, selfish & poaching, stabbing to death, evil, wrong, selfish\\
Hunters & poaching, stabbing to death, hunt game, hunt, hunting animals & poaching, hunt game, hunt, hunting animals, hunter & poaching, hunt game, hunt, hunting animals, hunter & poaching, stabbing to death, evil, people who exploit animals, unnatural thing & poaching, stabbing to death, evil, wrong, people who exploit animals\\
Local communities & poaching, stabbing to death, hunt game, hunt, hunting animals & poaching, hunt game, hunt, hunting animals, hunter & hunt game, hunt, poaching, hunter, hunting animals & stabbing to death, poaching, unnatural thing, evil, killing people & poaching, stabbing to death, evil, wrong, killing humans\\
Animal rights activists & stabbing to death, poaching, evil, unnatural thing, selfish & poaching, stabbing to death, evil, people who exploit animals, unnatural thing & stabbing to death, poaching, unnatural thing, evil, killing people & stabbing to death, people who exploit animals, poaching, cruelty, blood sport & stabbing to death, people who exploit animals, poaching, blood sport, unnatural thing\\
Environmentalists & poaching, stabbing to death, evil, wrong, selfish & poaching, stabbing to death, evil, wrong, people who exploit animals & poaching, stabbing to death, evil, wrong, killing humans & stabbing to death, people who exploit animals, poaching, blood sport, unnatural thing & stabbing to death, poaching, people who exploit animals, blood sport, unnatural thing\\
\midrule\midrule
Government officials & wild animal, living thing, control, animals, deer & control, living thing, pleasure, wild animal, animals & control, pleasure, living thing, kind, wild animal & hunt game, hunting animals, hunt, while hunting animals, hunter & hunt game, while hunting animals, hunting animals, hunter, hunt\\
Hunters & control, living thing, pleasure, wild animal, animals & control, pleasure, living thing, kind, joy & control, pleasure, living thing, kind, joy & hunt game, hunting animals, hunt, hunter, while hunting animals & hunt game, hunting animals, hunter, hunt, while hunting animals\\
Local communities & control, pleasure, living thing, kind, wild animal & control, pleasure, living thing, kind, joy & control, pleasure, kind, joy, living thing & hunt game, hunting animals, hunter, hunt, while hunting animals & hunt game, hunter, hunting animals, while hunting animals, hunt\\
Animal rights activists & hunt game, hunting animals, hunt, while hunting animals, hunter & hunt game, hunting animals, hunt, hunter, while hunting animals & hunt game, hunting animals, hunter, hunt, while hunting animals & goal, kind, many wild animals, endangered species, animals and sometimes people & goal, killing for food, outdoor activity, getting food, hunt\\
Environmentalists & hunt game, while hunting animals, hunting animals, hunter, hunt & hunt game, hunting animals, hunter, hunt, while hunting animals & hunt game, hunter, hunting animals, while hunting animals, hunt & goal, killing for food, outdoor activity, getting food, hunt & goal, killing for food, kind, good, getting food\\

\bottomrule
\end{tabular}
}
\caption{Agreement (top) and disagreement (bottom) concepts by Stakeholder for Animal Hunting.}
\label{app:tab:cs:concepts_by_stakeholder_AnimalHunting}
\end{table*}

\begin{table*}
\centering
\resizebox*{\linewidth}{!}{
\begin{tabular}{p{3cm}|p{3cm}p{3cm}p{3cm}p{3cm}}
\toprule
    & Advocacy groups for human rights & Criminals & Government & Victims and their families \\ \midrule
    Advocacy groups for human rights & murdering, committing crime, stupid, murder, injustice & committing crime, murdering, stupid, murder, reward & committing crime, stupid, murdering, murder, crimes & committing crime, stupid, murdering, murder, injustice\\
Criminals & committing crime, murdering, stupid, murder, reward & committing crime, murdering, stupid, reward, murder & committing crime, stupid, murdering, murder, crimes & committing crime, stupid, murdering, murder, crimes\\
Government & committing crime, stupid, murdering, murder, crimes & committing crime, stupid, murdering, murder, crimes & committing crime, crimes, someones who commits murder, stupid, murderers & committing crime, stupid, crimes, someones who commits murder, murdering\\
Victims and their families & committing crime, stupid, murdering, murder, injustice & committing crime, stupid, murdering, murder, crimes & committing crime, stupid, crimes, someones who commits murder, murdering & committing crime, someones who commits murder, stupid, crimes, murderers\\
\midrule\midrule
Advocacy groups for human rights & human right, rehab, human, change, imprisonment & human right, rehab, human, compassion, change & capital punishment, death penalty, death sentence, right to life, punishable by death & death penalty, death sentence, capital punishment, punishable by death, face death penalty\\
Criminals & human right, rehab, human, compassion, change & rehab, human, human right, feel remorse, compassion & capital punishment, right to life, death penalty, death sentence, punishable by death & capital punishment, death penalty, punishable by death, death sentence, execute\\
Government & capital punishment, death penalty, death sentence, right to life, punishable by death & capital punishment, right to life, death penalty, death sentence, punishable by death & kill, human killing, legal, change, deciding criminal s fate & human killing, meant as deterrent to crime, death, kill, die\\
Victims and their families & death penalty, death sentence, capital punishment, punishable by death, face death penalty & capital punishment, death penalty, punishable by death, death sentence, execute & human killing, meant as deterrent to crime, death, kill, die & kill, human killing, deciding criminal s fate, hanging, change\\

\bottomrule
\end{tabular}
}
\caption{Agreement (top) and disagreement (bottom) concepts by Stakeholder for Death Penalty.}
\label{app:tab:cs:concepts_by_stakeholder_DeathPenalty}
\end{table*}

\begin{table*}
\centering
\resizebox*{\linewidth}{!}{
\begin{tabular}{p{3cm}|p{3cm}p{3cm}p{3cm}p{3cm}}
\toprule
    & Parents & Students & School administrators & Teachers \\ \midrule
    Parents & pay off teacher, bank on failing in school, bribe, penalty, free & pay off teacher, bank on failing in school, bribe, penalty, hard work & pay off teacher, bank on failing in school, bribe, free, penalty & pay off teacher, bank on failing in school, bribe, free, penalty\\
Students & pay off teacher, bank on failing in school, bribe, penalty, hard work & bank on failing in school, pay off teacher, penalty, hard work, work hard & pay off teacher, bank on failing in school, bribe, penalty, free & pay off teacher, bank on failing in school, bribe, penalty, free\\
School administrators & pay off teacher, bank on failing in school, bribe, free, penalty & pay off teacher, bank on failing in school, bribe, penalty, free & pay off teacher, bank on failing in school, bribe, free, fee & pay off teacher, bank on failing in school, bribe, free, twenty bucks for every\\
Teachers & pay off teacher, bank on failing in school, bribe, free, penalty & pay off teacher, bank on failing in school, bribe, penalty, free & pay off teacher, bank on failing in school, bribe, free, twenty bucks for every & pay off teacher, bribe, bank on failing in school, twenty bucks for every, spend money\\
\midrule\midrule
Parents & get paid, reward, make money, money, feel good & get paid, reward, money, make money, twenty bucks for every & reward, value, make money, money, get paid & reward, make money, money, get paid, value\\
Students & get paid, reward, money, make money, twenty bucks for every & satisfaction, celebrate, twenty bucks for every, feel good, spend money & get paid, reward, make money, money, value & get paid, make money, reward, money, value\\
School administrators & reward, value, make money, money, get paid & get paid, reward, make money, money, value & education, better, work, school, make better world & education, better, work, school, make better world\\
Teachers & reward, make money, money, get paid, value & get paid, make money, reward, money, value & education, better, work, school, make better world & education, educate, learn lessons well, further education, get good grade\\

\bottomrule
\end{tabular}
}
\caption{Agreement (top) and disagreement (bottom) concepts by Stakeholder for Good Grades.}
\label{app:tab:cs:concepts_by_stakeholder_GoodGrades}
\end{table*}

\begin{table*}
\centering
\resizebox*{\linewidth}{!}{
\begin{tabular}{p{3cm}|p{3cm}p{3cm}p{3cm}p{3cm}}
\toprule
    & Illegal immigrants & Local communities & Employers & Government \\ \midrule
    Illegal immigrants & turn away, attack, exile, go home, roadblock & unnatural thing, stealing, criminals, government, attack & unnatural thing, stealing, government, criminals, attack & unnatural thing, stealing, criminals, attack, government\\
Local communities & unnatural thing, stealing, criminals, government, attack & unnatural thing, stealing, criminals, criminal act, illegal & stealing, unnatural thing, criminals, people who break laws, illegal & stealing, unnatural thing, criminals, criminal act, people who break laws\\
Employers & unnatural thing, stealing, government, criminals, attack & stealing, unnatural thing, criminals, people who break laws, illegal & stealing, people who break laws, criminals, unnatural thing, illegal & stealing, criminals, people who break laws, unnatural thing, illegal\\
Government & unnatural thing, stealing, criminals, attack, government & stealing, unnatural thing, criminals, criminal act, people who break laws & stealing, criminals, people who break laws, unnatural thing, illegal & criminals, stealing, people who break laws, unnatural thing, illegal\\
\midrule\midrule
Illegal immigrants & quality, justice, america, good feelings, change of location & amnesty, immigrants, deportations, immigrants people who, deport & immigrants, amnesty, immigrants people who, deportations, deport & amnesty, deportations, immigrants, immigrants people who, deport\\
Local communities & amnesty, immigrants, deportations, immigrants people who, deport & exile, turn away, order, ouster, return home & exile, order, turn away, law, ouster & exile, turn away, ouster, order, return home\\
Employers & immigrants, amnesty, immigrants people who, deportations, deport & exile, order, turn away, law, ouster & exile, order, law, country, citizen & exile, order, law, country, turn away\\
Government & amnesty, deportations, immigrants, immigrants people who, deport & exile, turn away, ouster, order, return home & exile, order, law, country, turn away & exile, law, order, turn away, immigration law\\

\bottomrule
\end{tabular}
}
\caption{Agreement (top) and disagreement (bottom) concepts by Stakeholder for Illegal Immigrants.}
\label{app:tab:cs:concepts_by_stakeholder_IllegalImmigrants}
\end{table*}

\begin{table*}
\centering
\resizebox*{\linewidth}{!}{
\begin{tabular}{p{3cm}|p{3cm}p{3cm}p{3cm}p{3cm}p{3cm}}
\toprule
    & School boards & School administrators & Teachers & Parents & Students \\ \midrule
    School boards & school uniform, uniform, bad, expressing yourself, reason & bad, uniform, school uniform, reason, expressing yourself & bad, uniform, school uniform, reason, expressing yourself & bad, very expensive, fashion, change, kids clothing & bad, very expensive, disguise, touchy about wearing uniforms, kids clothing\\
School administrators & bad, uniform, school uniform, reason, expressing yourself & bad, uniform, school uniform, reason, expressing yourself & bad, uniform, school uniform, reason, expressing yourself & bad, very expensive, disguise, fashion, change & bad, disguise, touchy about wearing uniforms, very expensive, kids clothing\\
Teachers & bad, uniform, school uniform, reason, expressing yourself & bad, uniform, school uniform, reason, expressing yourself & bad, uniform, school uniform, reason, expressing yourself & bad, very expensive, disguise, fashion, touchy about wearing uniforms & bad, disguise, touchy about wearing uniforms, pain, special way to dress\\
Parents & bad, very expensive, fashion, change, kids clothing & bad, very expensive, disguise, fashion, change & bad, very expensive, disguise, fashion, touchy about wearing uniforms & touchy about wearing uniforms, disguise, bad, special outfit, pain & disguise, touchy about wearing uniforms, bad, pain, special outfit\\
Students & bad, very expensive, disguise, touchy about wearing uniforms, kids clothing & bad, disguise, touchy about wearing uniforms, very expensive, kids clothing & bad, disguise, touchy about wearing uniforms, pain, special way to dress & disguise, touchy about wearing uniforms, bad, pain, special outfit & disguise, touchy about wearing uniforms, school uniform, uniform, required for schools to function effectively\\
\midrule\midrule
School boards & special way to dress, clothes, kids clothing, clothing, changing appearance & special way to dress, clothing, clothes, changing appearance, kids clothing & special way to dress, clothing, special outfit, clothes, changing appearance & school uniform, uniform, required for schools to function effectively, school, improving image & school uniform, uniform, required for schools to function effectively, school, ideal\\
School administrators & special way to dress, clothing, clothes, changing appearance, kids clothing & special way to dress, clothing, changing appearance, special outfit, clothes & special outfit, special way to dress, clothing, changing appearance, organization & school uniform, uniform, required for schools to function effectively, school, ideal & uniform, school uniform, required for schools to function effectively, school, ideal\\
Teachers & special way to dress, clothing, special outfit, clothes, changing appearance & special outfit, special way to dress, clothing, changing appearance, organization & special outfit, special way to dress, clothing, changing appearance, like & school uniform, uniform, required for schools to function effectively, school, ideal & uniform, school uniform, required for schools to function effectively, school, ideal\\
Parents & school uniform, uniform, required for schools to function effectively, school, improving image & school uniform, uniform, required for schools to function effectively, school, ideal & school uniform, uniform, required for schools to function effectively, school, ideal & expressing yourself, change, dress themselves, motivation, reason & expressing yourself, dress themselves, change, motivation, reason\\
Students & school uniform, uniform, required for schools to function effectively, school, ideal & uniform, school uniform, required for schools to function effectively, school, ideal & uniform, school uniform, required for schools to function effectively, school, ideal & expressing yourself, dress themselves, change, motivation, reason & change, expressing yourself, fashion, dress themselves, student\\

\bottomrule
\end{tabular}
}
\caption{Agreement (top) and disagreement (bottom) concepts by Stakeholder for School Uniforms.}
\label{app:tab:cs:concepts_by_stakeholder_SchoolUniforms}
\end{table*}

\section{Usage of AI assistants}
We use GitHub Copilot (\url{https://github.com/features/copilot}) for speeding up programming, and ChatGPT (\url{https://chat.openai.com}) to aid with reformulations. The content of this work is our own, and not largely inspired by AI assistants.

\begin{acronym}
    \acro{FT-LM}[FT-LM]{\textbf{F}ine-\textbf{T}uned \textbf{L}anguage \textbf{M}odel}
    \acro{LLM}[LLM]{\textbf{L}arge \textbf{L}anguage \textbf{M}odel}
    \acro{PI-LM}[PI-LM]{\textbf{P}rompt-\textbf{I}nstructed \textbf{L}anguage \textbf{M}odel}
    \acro{PSV}[PSV]{\textbf{P}erspectivized \textbf{S}tance \textbf{V}ector}
    \acro{CBR}[CBR]{\textbf{C}ase \textbf{B}ased \textbf{R}easoning}
\end{acronym}

\end{document}